\documentclass[10pt,twocolumn,letterpaper]{article}

\usepackage[pagenumbers]{cvpr} 

\usepackage{algorithm}
\usepackage{algorithmic}
\usepackage{booktabs}
\usepackage{colortbl}
\usepackage{xcolor}
\usepackage{multirow}
\usepackage{amsmath}
\usepackage{mathtools}
\usepackage{kotex}
\definecolor{slotrow}{HTML}{F4DCC7}  
\newcommand{\cmark}{\ding{51}}%
\newcommand{\xmark}{\ding{55}}%
\usepackage{pifont}
\usepackage{soul} 
\usepackage{amsfonts}
\usepackage{subcaption}
\usepackage{wrapfig}
\usepackage[accsupp]{axessibility}  

\usepackage{bbm}
\usepackage{boldline}
\definecolor{cvprblue}{rgb}{0.21,0.49,0.74}
\usepackage[pagebackref,breaklinks,colorlinks,allcolors=cvprblue]{hyperref}

\def\paperID{11955} 
\def\confName{CVPR}
\def\confYear{2026}

\title{Reconstruction-Guided Slot Curriculum: \\ Addressing Object Over-Fragmentation in Video Object-Centric Learning}

\author{
    WonJun Moon \;\;\; Hyun Seok Seong \;\;\; Jae-Pil Heo\thanks{Corresponding Author} \\
    Sungkyunkwan University, South Korea \\
{\tt\small \{wjun0830, gustjrdl95, jaepilheo\}@skku.edu}
}

\begin{document}
\maketitle
\begin{abstract}
Video Object‑Centric Learning seeks to decompose raw videos into a small set of object slots, but existing slot‑attention models often suffer from severe over‑fragmentation.
This is because the model is implicitly encouraged to occupy all slots to minimize the reconstruction objective, thereby representing a single object with multiple redundant slots.
We tackle this limitation with a reconstruction‑guided slot curriculum~(SlotCurri). 
Training starts with only a few coarse slots and progressively allocates new slots where reconstruction error remains high, thus expanding capacity only where it is needed and preventing fragmentation from the outset. 
Yet, during slot expansion, meaningful sub‑parts can emerge only if coarse‑level semantics are already well separated; however, with a small initial slot budget and an MSE objective, semantic boundaries remain blurry.
Therefore, we augment MSE with a structure‑aware loss that preserves local contrast and edge information to encourage each slot to sharpen its semantic boundaries.
Lastly, we propose a cyclic inference that rolls slots forward and then backward through the frame sequence, producing temporally consistent object representations even in the earliest frames.
All combined, SlotCurri addresses object over-fragmentation by allocating representational capacity where reconstruction fails, further enhanced by structural cues and cyclic inference.
Notable FG-ARI gains of +6.8 on YouTube-VIS and +8.3 on MOVi-C validate the effectiveness of SlotCurri.
Our code is available at \href{github.com/wjun0830/SlotCurri}{github.com/wjun0830/SlotCurri}.
\end{abstract}    
\section{Introduction}
\label{introduction}

Video Object-Centric Learning~(VOCL) aims to decompose raw videos into compact object slots. 
By transforming high-dimensional spatio-temporal features into a structured latent space, object-centric models can assist in effectively capturing spatial relationships and temporal dynamics at the object level~\cite{slotvlm}. 
This slot-based representation provides a robust foundation for diverse downstream tasks, including scene understanding~\cite{slotvae} and video segmentation~\cite{guidedslot}.

\begin{figure}[t]
\centering
\vspace{-0.3cm}
\includegraphics[width=0.89\columnwidth]{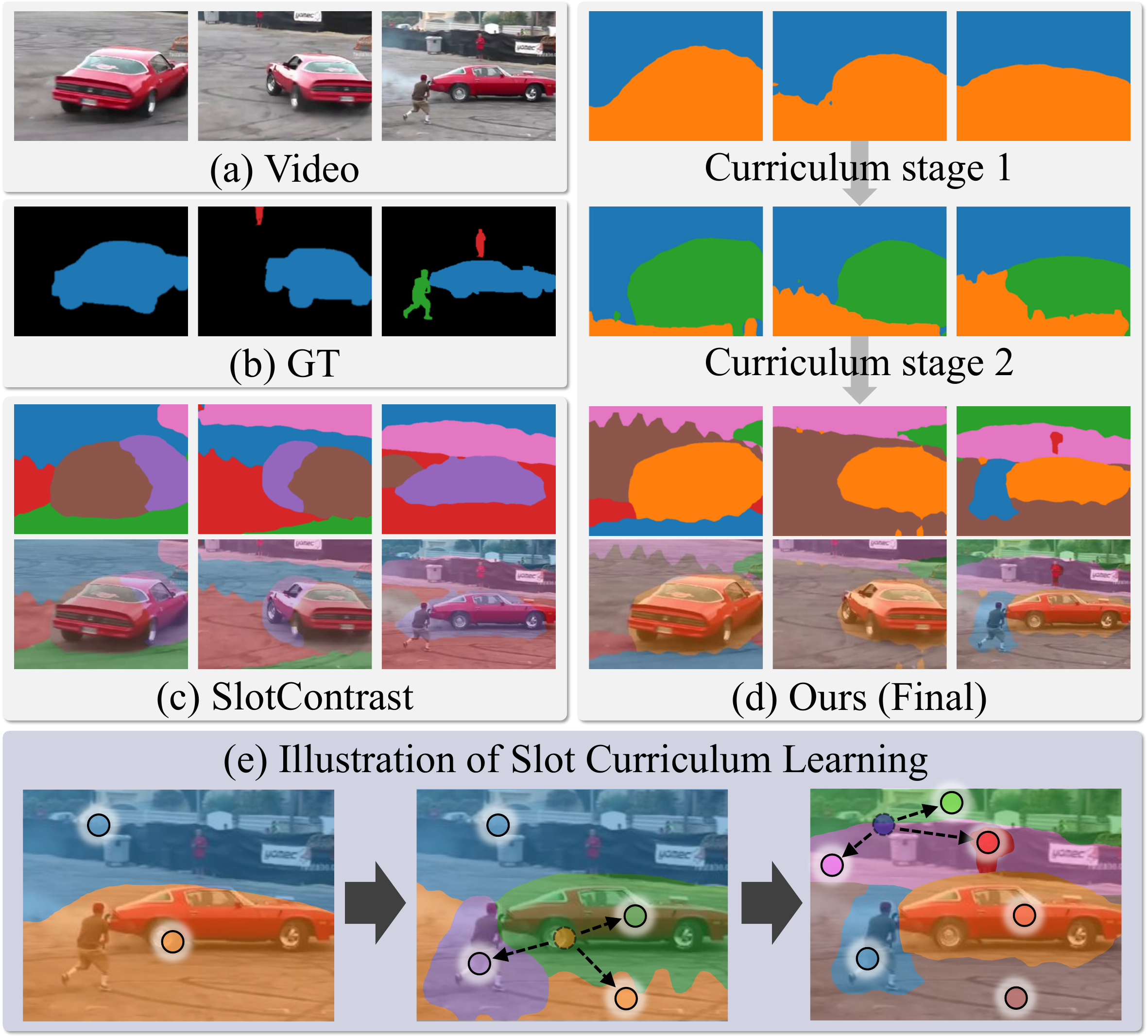} 
\vspace{-0.25cm}
\caption{
\textbf{Motivation and method overview.}
(a)~Original video frames. 
(b)~Ground-truth masks. 
(c)~Our baseline SlotContrast~\cite{slotcontrast} learns to decompose with a high slot budget from scratch.
This approach often results in over-fragmentation, splitting a single object~(car) across multiple slots~(brown and purple slots).
(d)~In contrast, our curriculum learning significantly reduces this over-fragmentation, yielding semantically coherent slots.
(e)~This is achieved by starting with a minimal slot budget~(\textit{e.g.}, 2 slots) and progressively spawning new slots in regions that existing slots fail to capture, thereby preserving object-level slots.
}
\label{fig1.motiv}
\vspace{-0.45cm}
\end{figure}

However, existing slot-based models encounter critical limitations in the absence of supervision regarding object scales, shapes, and counts. 
To be specific, the model is implicitly pressured to fully utilize all available slots since reconstruction quality generally improves as the slot budget grows~\cite{adaptiveslot}.
As a consequence, the model often exhibits over-fragmentation, splitting a single object across several slots. 
This not only introduces unnecessary redundancy but also means that individual slots fail to capture a complete or precise representation of the object.
Fig.~\ref{fig1.motiv} illustrates how the previous state-of-the-art approach tends to encode a single object across multiple slots, undermining the interpretability and effectiveness of object-centric representations.

In this paper, we propose SlotCurri, which treats the number of available slots as a curriculum variable, progressively increasing the granularity of representation throughout training. 
We begin training with a minimal set of coarse slots~(\textit{e.g.}, 2 slots), allowing the model to initially focus on broad spatial grouping of pixels. 
Once learning stabilizes, we gradually expand the number of slots in distinct stages. 
Crucially, newborn slots are initialized based on reconstruction loss, selectively duplicating slots exhibiting higher error rates and perturbing them with carefully scaled noise. 
This noise is calibrated such that each newborn slot inherits its parent’s representation and is assigned to a sub-region with high reconstruction error, enabling it to capture distinct yet related parts without drifting far from its parent.

In addition, current VOCL models are typically trained with mean-squared-error~(MSE) reconstruction loss. 
Since the MSE treats every pixel independently and minimizes the averaged error, it inevitably blurs spatial details in decoded representations and smears the true boundaries between objects~\cite{mseblur1, mseblur2}. 
The problem is exacerbated when only a few slots are available during the early stages of our curriculum learning.
When training begins with only a small number of slots, each slot is forced to cover a very large and semantically diverse region of the scene, which makes the borders between entities hard to disentangle.
As a result, features from neighboring objects and background regions intermix, blurring slot boundaries and making it unclear which slot corresponds to which object.
To address this, we employ a structure-aware loss that complements MSE by explicitly preserving local contrast and edge information. 
Enforcing structural cues during reconstruction trains each slot to form sharper boundaries, which simplifies identity separation when additional slots are introduced.

Lastly, although our SlotCurri significantly reduces over‑fragmentation, the earliest frames may still be relatively under-fitted since contextual cues cannot be leveraged.
To balance slot quality over time, we introduce a cyclic inference: slots are first propagated forward to the last frame and then cycled backward to the first, producing stable and consistent encodings across the entire sequence.

Our main contributions are summarized as follows:
\begin{itemize}
    \item We introduce SlotCurri, a reconstruction-guided slot curriculum, progressively spawning new slots in regions of high reconstruction error.
    \item We exploit a structure-aware loss to stabilize the coarse-to-fine decomposition by preserving local structures.
    \item We introduce a cyclic inference strategy to leverage aggregated contextual cues at earlier frames and enhance object consistency with negligible extra cost.
\end{itemize}
Taken together, these components substantially reduce object over-fragmentation and deliver state-of-the-art performance on YouTube-VIS, MOVi-C, and MOVi-E.



\section{Related Work}
\label{relatedwork}
\subsection{Object‑Centric Representation Learning.}
Object-centric learning unsupervisedly groups perceptual inputs into distinct object entities without supervision, mimicking human scene understanding~\cite{greff2016tagger,OCL1,OCL2,OCL3,OCL4,OCL5,kakogeorgiou2024spot,zhao2025slot,srl}.
Among recent methods, Slot Attention~\cite{slotattention} has emerged as a simple yet powerful mechanism that assigns latent slots to coherent objects.  
Its real‑world applicability has been demonstrated in natural images~\cite{seitzer2023bridging} and further extended to downstream tasks such as unsupervised segmentation~\cite{groupvit_OCL_slot_seg,slotvps}, retrieval~\cite{slot_for_retrieval1,slot_for_retrieval2}, question answering~\cite{slotvlm} and generation~\cite{jiang2023object}.

Recently, adapting slots to the video domain is being spotlighted~\cite{steve, slotvlm}.
Early attempts, such as SAVi~\cite{savi}, leverage a bounding box cue in the first frame or sparse depth from LiDAR to anchor slot identities in driving scenes.  
SOLV~\cite{aydemir2023self} adopts a masked‑autoencoder objective and introduces slot‑merging to overcome the over-fragmentation.
VideoSAUR~\cite{videosaur} models explicit patch‑level motion, predicting future feature similarities to bind slots across frames.
SlotContrast~\cite{slotcontrast} shows that contrastive learning can further improve temporal consistency: it forms positive pairs between the same slot in consecutive frames and contrasts them against other slots in the batch. 
Similar to SOLV~\cite{aydemir2023self}, we target the over-fragmentation problem; yet, instead of first over‑producing slots and then merging them, our strategy eliminates fragmentation a priori by progressively adding slots only in regions with persistently high reconstruction error.
This encourages more stable and semantically aligned slots, as merging may fail once contrastive pressure has pushed slots to encode distinct representations even when their constituent patches share similar semantics.

\subsection{Curriculum Learning}
Curriculum learning was originally proposed to imitate the way humans learn; the model is given easier samples at earlier steps and gets exposed to more difficult samples as training progresses~\cite{bengio2009curriculum}.
Since then, it has been introduced for various downstream tasks, including image classification~\cite{curriculum_classification}, object detection~\cite{curriculum_detection}, long-tailed recognition~\cite{curriculum_longtail}, and retrieval-augmented generation~\cite{curriculum_rag}.
Our work shares the motivation with these works in that we aim to establish a solid foundation and gradually increase the learning capacity.
However, our strategy is tailored for VOCL in that we treat the number of slots as a curriculum variable; in earlier training steps, we learn to partition coarse-level semantics and gradually adapt slots to encode each entity.
\section{Method}
\label{sec:method}

\subsection{Preliminaries}

\begin{figure}[t]
\centering
\vspace{-0.25cm}
\includegraphics[width=0.85\columnwidth]{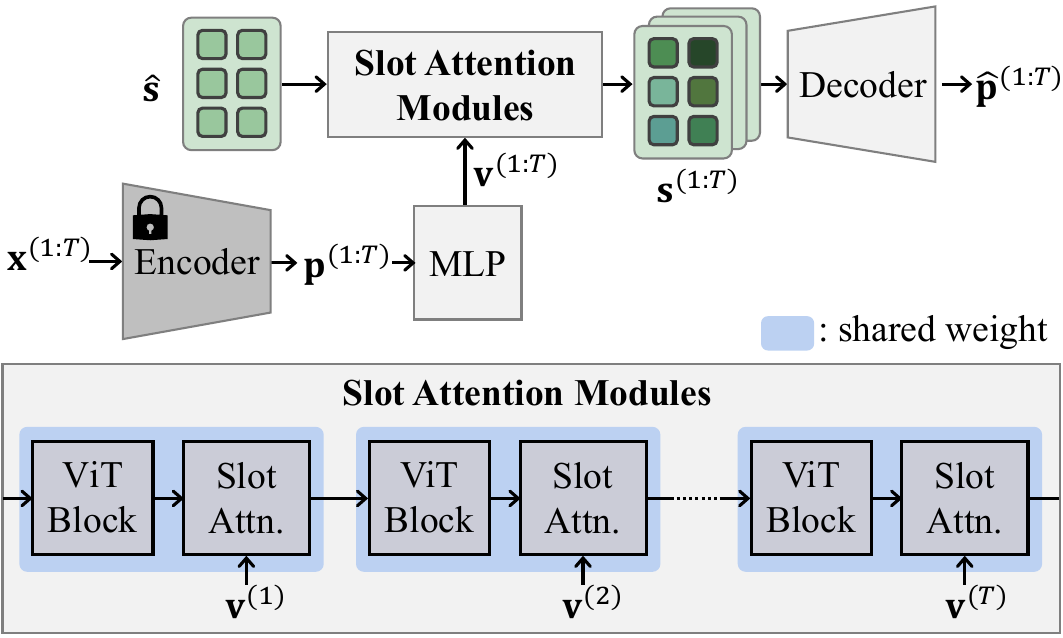} 
\vspace{-0.15cm}
\caption{
\textbf{An architectural overview of our framework.}
}
\label{Fig.overview}
\vspace{-0.45cm}
\end{figure}

\textbf{Video Object-Centric Learning.}
As illustrated in Fig.~\ref{Fig.overview}, given an unlabeled video $\mathbf{x}^{(1:T)}$ of length $T$, a VOCL pipeline first converts each frame $\mathbf{x}^{(t)}$~($t=1,\dots,T$) into a patch-level representation $\mathbf{p}^{(t)}$ using a vision foundation model~(\textit{e.g.}, DINO‑v2~\cite{Dinov2}).
These are then passed through a task-specific MLP layer to produce frame representations $\mathbf{v}^{(t)}$.
Finally, a slot‑attention~\cite{slotattention} module decomposes it into a set of $K$ latent object slots $\mathbf{s}^{(t)} = \left\{ \mathbf{s}^{(t,k)} \in \mathbb{R}^{D} \;\middle|\; k=1,\dots,K \right\}$ using globally shared slot placeholders $\hat{\mathbf{s}} \in \mathbb{R}^{K \times D}$
where each slot is intended to capture a single scene constituent. 
We note that the slot placeholders $\hat{\mathbf{s}}$ are recurrently refined in a sequential manner from $t\!=\!1$ to $T$, producing the frame-specific slot features $\mathbf{s}\in \mathbb{R}^{T\times K \times D}$.
The module refines these slots for $L$ iterations through a sequence of projection, multi‑head attention, and Gated Recurrent Unit~\cite{gru}, implementing a soft Expectation–Maximization~(EM)~\cite{dempster1977maximum}. 
Finally, a decoder maps the refined slots back to the reconstructed patch $\hat{\mathbf{p}}^{(t)}$, encouraging each slot to explain a coherent image region, thus promoting an object‑centric partition of the frame.


\textbf{SlotContrast.}
Our method builds on SlotContrast~\cite{slotcontrast}, which enforces temporal consistency within each slot. 
Specifically, SlotContrast introduces a slot‑slot contrastive loss that maximizes the similarity of a given slot across consecutive time‑steps while simultaneously repelling all other slots in the mini-batch.
We note that contrastive pairs are formed only with the successive frame.
To formally define $\mathcal{L}_{\text{SSC}}$, we extend the slot representation to $\mathbf{s} \in \mathbb{R}^{B \times T \times K \times D}$, explicitly including the batch dimension.
Slot-slot contrastive loss for the $b$-th video is expressed as $\mathcal{L}_{\text{SSC}} = \frac{1}{(T-1) K}\sum_{i=2}^{T}\sum_{j=1}^{K} \ell_{\text{ssc}}^{(i,j)}$ and $\ell_{\text{ssc}}^{(i,j)}=$
\begin{equation}
\label{Eq.slotslotcontrastiveloss}
     \!\!\!\!\!-\!\log\!
     \frac{\exp\!\bigl(\operatorname{sim}(\mathbf{s}^{(b,i-1,j)},\mathbf{s}^{(b,i,j)})/\tau\bigr)}
          {\displaystyle
           \!\!\sum_{c=1}^{B}\!\sum_{k=1}^{K}
           \mathbf{1}_{[(c,k)\neq (b,j)]}\!
           \exp\!\bigl(\operatorname{sim}(\mathbf{s}^{(b,i-1,j)},\!\mathbf{s}^{(c,i,k)})/\tau\bigr)}\!,
\end{equation}
where $\operatorname{sim}(\cdot,\cdot)$ is cosine similarity, $\tau$ is a scaling parameter, and $\mathbf{1}_{[(c,k)\neq (b,j)]}$ is an indicator function.
Overall, the loss of SlotContrast is defined as the sum of the reconstruction loss~(\(\mathcal{L}_{\text{MSE}}\)) computed between the backbone feature \(\mathbf{p}\) and its reconstruction \(\hat{\mathbf{p}}\), and the slot‑to‑slot contrastive loss~(\(\mathcal{L}_{\text{SSC}}\)).
For the rest of the paper, we omit $b$ for brevity and denote the $k$-th slot at $t$-th frame simply as $\mathbf{s}^{(t,k)}$.

\begin{figure*}[t]
\centering
\vspace{-0.27cm}
\includegraphics[width=0.9\textwidth]{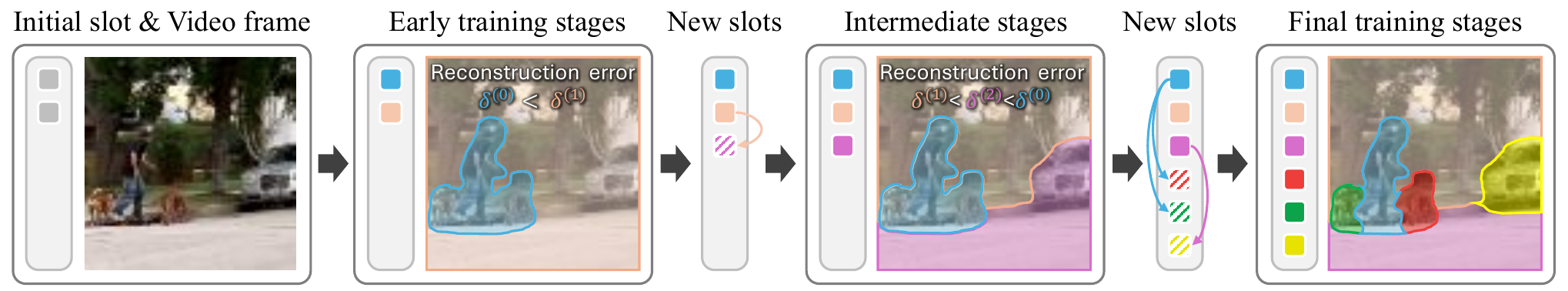} 
\vspace{-0.22cm}
\caption{
\textbf{Illustration of our reconstruction‑guided slot curriculum learning.}
The model begins with a few coarse slots that coarsely segment large regions.
At each scheduled iteration, slots whose reconstruction errors~($\delta$) remain high are duplicated and perturbed with distance‑aware noise~(\cref{eq.distance-aware-noise}), creating new slots that focus on the under‑explained areas.
Iterating this process across training stages gradually enlarges the slot set, yielding a fine‑grained partition with clearly separated object regions.
}
\label{Fig.curriculum}
\vspace{-0.37cm}
\end{figure*}

\subsection{Motivation}
\label{sec:scheduling}
When a model begins training with the maximum slot budget already available, the soft‑EM dynamics of Slot Attention have no explicit pressure to assign one object to one slot. 
In other words, as long as the reconstruction loss is reduced, several slots may cooperate on the same object~\cite{adaptiveslot}, triggering an over‑fragmentation problem.

This over-fragmentation poses critical practical challenges.
In downstream tasks such as reasoning, tracking, and video summarization~\cite{jiang2024compositional,jiang2019scalor,slotvlm}, slots are expected to correspond finely to individual objects so that their identities and dynamics can be extracted with minimal effort. 
Yet, when a single object is split across slots, not only does the interpretability of the pipeline suffer, but also the redundant and fragmented slots reduce computational efficiency.


To mitigate over-fragmentation, we propose slot curriculum learning, which starts with a minimal slot budget and gradually expands it as training progresses. 
With only a few slots available at early stages, each slot is forced to cover broad, semantically coarse regions.
Once new slots are introduced, this setup naturally induces a coarse-to-fine partitioning.
Specifically, patches that already form coherent semantic groups are tightly bound to their existing slots, and thus remain stable, whereas the components in coarsely grouped, semantically mixed regions are only weakly bound to their slots and are thus more likely to be detached and assigned to the newly spawned slots.
This process allows the overall slot set to refine toward semantically coherent entities.
In \cref{sec.simple}, we introduce a simple curriculum that implicitly facilitates this partitioning, whereas in \cref{sec:slotcurriculum}, we propose a method that explicitly guides new slots to focus on the regions of high reconstruction error, which typically correspond to coarsely grouped, mixed-object regions.

\subsection{Simple Slot Curriculum Learning}
\label{sec.simple}
One simple way to implement a curriculum baseline is to initialize new slots randomly, following prior works~\cite{slotcontrast,slotvps}. 
Simply put, we start training with a small number of slots, incrementally adding more randomly initialized slots at predefined intervals. 
Formally, let $K_{\text{init}}$ denote the initial number of slots.
Given $M$ curriculum stages, we design an accelerated slot schedule; the slot budget $K^{(m)}$ at stage $m$~($m=0, \dots, M-1$) increases with an accelerated rule controlled by a base-increment parameter $\sigma$:
\begin{equation}
\label{eq.slotnum}
    K^{(m)} = K_{\text{init}} + m \cdot \sigma + 3m(m-1) / 2,
\end{equation}
where $K_{\text{init}}$ is set to 2 in our work.
In short, training starts with $K_{\text{init}}$ slots, and at each predefined iteration, the slot budget is increased to $K^{(m)}$.
As demonstrated in our ablation study~(Tab.~\ref{tab:ablation}), even this simple curriculum learning boosts performance by forcing slots to first capture coarse semantic features. 
We attribute this performance improvement to reduced over-fragmentation, as slots that have learned coherent semantic regions become tightly bound to their associated patches, discouraging arbitrary splits.

\subsection{SlotCurri: Reconstruction-Guided \\ Slot Curriculum Learning}
\label{sec:slotcurriculum}
Despite its simplicity and effectiveness, simple slot curriculum learning can be further improved by a more informed initialization of newly spawned slots.
This is because randomly initialized slots often miss underrepresented regions, wasting capacity on areas already well modeled.
Therefore, instead of expanding slot capacity without regard to where it is needed, we propose to strategically direct new slots toward regions that are most challenging for current slots to represent, as illustrated in Fig.~\ref{Fig.curriculum}.
Specifically, we identify the slots with the highest reconstruction error and spawn child slots by duplicating them.
Each child is initialized by adding Gaussian noise, scaled proportionally to the parent's nearest-neighbor distance to other slots in the feature space.
This reconstruction-guided curriculum preserves well-modeled objects while prioritizing computational resources for areas that demand improved representation.

To locate new slots in under-explained regions, we measure how much reconstruction error each slot is responsible for.
We compute the slot-wise error $\delta$ by weighting the MSE loss at each pixel with the slot's reconstruction weights $\alpha$ as:
\begin{equation}
\delta^{(k)} = \sum_{t,h,w}\alpha^{(k,t,h,w)}\cdot\mathcal{L}_{\text{MSE}}^{(t,h,w)},
\end{equation}
where $\alpha^{(k,t,h,w)}$ denotes the decoding weight for $k$-th slot at spatio-temporal location $(t,h,w)$, and $\mathcal{L}_{\text{MSE}}^{(t,h,w)}$ represents the corresponding pixel-wise reconstruction loss.
Using this reconstruction-error-driven $\delta$ as the slot-spawning criterion naturally suppresses idle slots from being further partitioned, since such slots have near-zero error mass.

Then, at the transition from curriculum stage $m\!-\!1$ to $m$~(with $K^{(m-1)}$ current slots), we convert the slot-wise reconstruction error $\delta^{(k)}$ into non-negative weights:
\begin{equation}
    w^{(k)} \;=\;\frac{\delta^{(k)}}{\sum_{j=1}^{K^{(m-1)}} \delta^{(j)}}, 
    \qquad \sum\nolimits_{k=1}^{K^{(m-1)}} w^{(k)}=1.
\end{equation}
These are then used to determine how many replicas are assigned to each slot.
At stage $m$, we add $N_{\text{new}}^{(m)} = K^{(m)} - K^{(m-1)}$ new slots.
Given these $N_{\text{new}}^{(m)}$ slots to distribute, we first compute the fractional allotments $\tilde n^{(k)} = w^{(k)}\,N_{\text{new}}^{(m)}$.
We convert these to integer replica counts by computing the floor of each allocation and the number of remaining slots:
\begin{equation}
n^{(k)} = \bigl\lfloor \tilde n^{(k)} \bigr\rfloor,
\quad
r = N_{\text{new}}^{(m)} - \sum\nolimits_{k=1}^{K^{(m-1)}} n^{(k)}.
\end{equation}
The remaining $r$ slots are distributed, one each, to the $r$ slots exhibiting the largest fractional residues ($\tilde n^{(k)} - n^{(k)}$).
This deterministic rounding guarantees $\sum_k n^{(k)} = N_{\text{new}}^{(m)}$ while prioritizing slots with higher reconstruction errors.

Yet, duplicating global slot placeholders $\hat{\mathbf{s}}$ may just split an already well-captured object into smaller pieces, aggravating over-fragmentation.
To ensure that new slots explore previously unexplored regions rather than duplicating existing slot identities, we initialize replicas by perturbing parent slot embeddings. 
Specifically, each newborn slot is created by perturbing its parent slot with a random unit vector, whose magnitude is proportional to the parent’s distance to its most similar neighbor and its relative feature norm:
\begin{equation}
\label{eq.distance-aware-noise}
\hat{\mathbf{s}}^{(k^*)} = \hat{\mathbf{s}}^{(k)} + \beta \cdot d_{\text{nearest}}^{(k)} \cdot \left( \frac{\| \hat{\mathbf{s}}^{(k)} \|}{\mu_{\text{norm}}} \right) \cdot \mathbf{v}
\end{equation}
where $k \in \{1, \dots, K^{(m-1)}\}$ indexes the parent slots, $k^* \in \{K^{(m-1)}+1, \cdots, K^{(m-1)}+N_\text{new}^{(m)}\}$ indexes the newly created child slots, $\mathbf{v}$ is a random unit vector sampled from a unit sphere to provide direction and $d_{\text{nearest}}^{(k)}$ is the Euclidean distance from slot $k$ to its closest slot placeholder. 
The term $\| \hat{\mathbf{s}}^{(k)} \| / \mu_{\text{norm}}$ is an adaptive scaling factor where $\mu_{\text{norm}}$ is the average L2-norm of all current slot placeholders, ensuring the noise magnitude remains relative to the feature scale. 
Finally, $\beta$ is a hyperparameter controlling the overall perturbation strength.
This initialization strategy guides newly spawned slots toward previously underrepresented or poorly reconstructed regions by allocating more child slots to those with higher reconstruction loss.
As a result, the expanded slot budget is utilized more effectively, reducing redundancy and mitigating over-fragmentation.

\begin{figure}[t]
\centering
\vspace{-0.3cm}
\includegraphics[width=0.9\columnwidth]{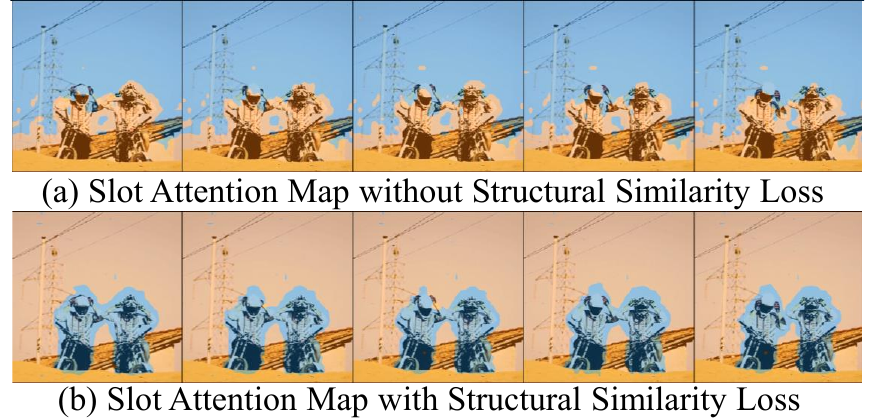} 
\vspace{-0.15cm}
\caption{
\textbf{Slot attention maps at an early training} (Curriculum stage 1 with 2 slots). 
(a) Without structural loss, slots broadly capture coarse semantic divisions but exhibit blurred boundaries and partial fragmentation of object parts~(helmet of person on the left), indicating weak structural consistency.
(b) With structural loss, slots form clearer, sharper boundaries and more coherent object representations, significantly improving object grouping quality.
}
\label{Fig.SSIMLoss}
\vspace{-0.3cm}
\end{figure}

\noindent\textbf{Structure-Aware Reconstruction Loss.}
\label{sec:structure-loss}
Current VOCL models are typically trained using mean-squared-error~(MSE) reconstruction loss, which measures errors independently per pixel. 
While convenient and efficient, MSE inherently promotes averaged predictions, thus blurring spatial details and obscuring the true boundaries between objects~\cite{mseblur1, mseblur2}. 
This issue is further exacerbated during early curriculum stages when only a few slots are available. 
In such situations, each slot is forced to represent large, diverse regions, causing background and adjacent object features to overlap significantly. 
This overlap makes it challenging for slots to form distinct and coherent object identities.
For example, in Fig.~\ref{Fig.SSIMLoss}~(a), we observe that the same entity is attended by two distinct slots~(whether the person on the left or his/her helmet is regarded as the primary entity).
Likewise, we observe that relying solely on the MSE objective may cause a single object to be independently partitioned, in which the error may propagate over curriculum stages.


To mitigate these drawbacks, we employ a Structural Similarity~(SSIM) loss~\cite{ssim} that explicitly preserves local structural cues within the reconstructed features, complementing the pixel-wise MSE loss. 
Specifically, SSIM measures the similarity between two signals as:
\begin{equation}
\label{eq.ssim}
\operatorname{SSIM}(\mathbf{x},\mathbf{y})
=\frac{(2\mu_x\mu_y+c_1)(2\sigma_{xy}+c_2)}
       {(\mu_x^{2}+\mu_y^{2}+c_1)(\sigma_x^{2}+\sigma_y^{2}+c_2)},
\end{equation}
where $\mu_\bullet$, $\sigma_\bullet^{2}$, and $\sigma_{xy}$ are the mean, variance, and cross-covariance computed from each sliding cubic window, responsible for evaluating luminance, contrast, and structural similarities. 
Constants $c_1$ and $c_2$ serve as regularizers.


We compute SSIM on spatio-temporal cubes of size $3{\times}3{\times}3$, sliding across the $(T,H,W)$ grid of decoded patches to assess both the spatial and temporal coherence.
Formally, the resulting SSIM loss for each sample is defined as the average of channel-wise SSIM across all windows $\Omega$: 
\begin{equation}
\mathcal{L}_{\text{SSIM3D}}
      = 1 - \frac{1}{|\Omega|}\sum\nolimits_{u\in\Omega}
        \operatorname{SSIM}\bigl(\hat{\mathbf{p}}^{(u)},\mathbf{p}^{(u)}\bigr),
\end{equation}
where $\mathbf{p}$ and $\hat{\mathbf{p}}$ denote ground-truth~(GT) and reconstructed patch features, respectively.
Note that we simplify the cube indexing with $u$ that indexes a valid $3{\times}3{\times}3$ cube within $\mathbf{p}$ and $\hat{\mathbf{p}}$, and $\mathbf{p}^{(u)}$ refers to the corresponding sub-volume extracted from $\mathbf{p}$.
By complementing the MSE-based reconstruction objective with this structural constraint, each slot learns sharper and more distinct boundaries, as observed in Fig.~\ref{Fig.SSIMLoss}~(b). 
Consequently, the subsequent slot expansion operates on already coherent
regions, enabling a principled coarse-to-fine partitioning of the
scene rather than uncontrolled over-fragmentation.
Our final loss is formulated as:
$
\mathcal{L} = \mathcal{L}_{\text{MSE}} + \lambda_{\text{SSC}}\mathcal{L}_{\text{SSC}} + \lambda_{\text{SSIM3D}}\mathcal{L}_{\text{SSIM3D}}.
$

\begin{figure}[t]
\centering
\vspace{-0.3cm}
\includegraphics[width=0.88\columnwidth]{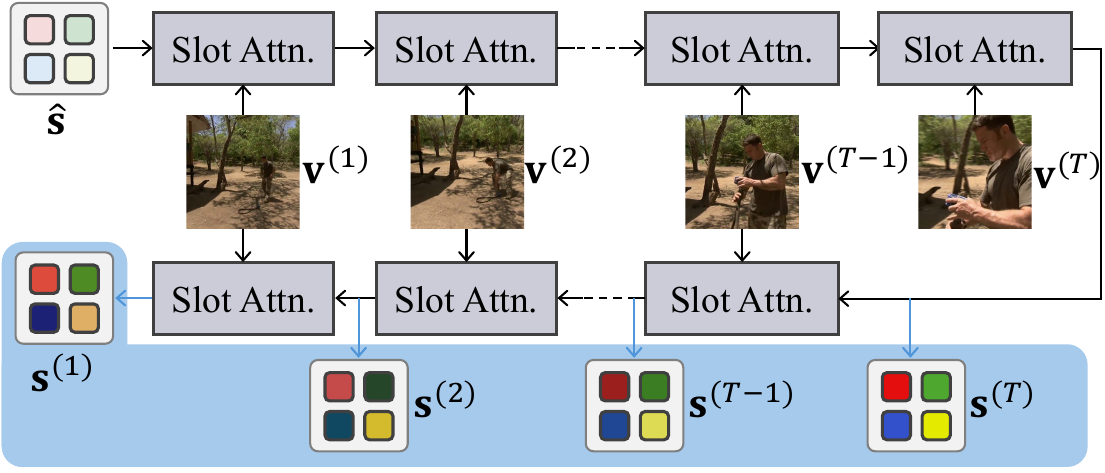} 
\vspace{-0.18cm}
\caption{
\textbf{Illustration of cyclic inference.} 
Slots on blue background are the output set preserving contextual information.
}
\label{Fig.cycleinference}
\vspace{-0.38cm}
\end{figure}

\subsection{Cyclic Inference for Temporal Consistency}
\label{sec:cyclic-inference}
While our slot curriculum learning effectively alleviates slot over-fragmentation, we find that the earliest video frames often remain under-fitted compared to later frames, due to the limited accumulated contextual information. 
To improve temporal consistency and achieve balanced reconstruction quality across the video sequence, we propose a cyclic strategy applied only during inference, as shown in Fig.~\ref{Fig.cycleinference}.

Specifically, we first perform forward propagation, sequentially updating slots from the initial frame to the last frame while accumulating temporal context. 
Subsequently, we reverse the process through backward propagation, updating slots from the last frame back to the first frame. 
The slot representations from this latter backward propagation are then used for final mask decoding.
This cyclic approach ensures that slot encodings incorporate both past and future contextual information, resulting in more temporally consistent and robust object slots throughout the video sequence.
We note that cyclic inference introduces only minimal computational overhead, increasing inference runtime by merely 0.3\%; the average inference time over five runs on YouTube-VIS increases slightly from 286s to 287s. 
This minimal cost implies that the slot attention module is computationally lightweight compared to the much heavier encoder and decoder stages. 
Given this efficiency, we expect that cyclic inference can be easily integrated into streaming applications; we validate a chunk-wise variant in Appendix.






\section{Experiments}
\subsection{Experimental Details}
\noindent\textbf{Experimental Settings.} We use real-world large-scale YouTube-VIS 2021~\cite{ytvis,ytvis2021_19,ytvis2021_21} and the synthetic MOVi-C and MOVi-E datasets for evaluation~\cite{greff2022kubric}.
For assessment, we use the video foreground adjusted rand index~(FG-ARI)~\cite{rand1971objective, fgari} and mean best overlap~(MBO)~\cite{seitzer2023bridging}.
FG‑ARI measures the clustering similarity of each foreground pixel.
It treats its GT instance label and the predicted slot assignment as cluster labels, and is invariant to any permutation of the slot indices.
On the other hand, MBO evaluates mask quality by computing, for each GT object mask, the maximum intersection‑over‑union with any predicted slot mask.

\noindent\textbf{Implementation Details.}
Following SlotContrast~\cite{slotcontrast}, we adopt DINOv2~\cite{Dinov2} as the visual backbone, instantiating ViT‑B/14 for YouTube‑VIS and MOVi‑E, and ViT‑S/14 for MOVi‑C.  
We set the number of curriculum stages $M$ to 3, scheduling slot expansions at 10\% and 25\% of the total training iterations.
The base-increment hyperparameter $\sigma$~(defined in \cref{eq.slotnum}) is set to 1, 3, and 5 for the YouTube-VIS, MOVi-C, and MOVi-E datasets, respectively.
This ensures that our final maximum slot count, $K^{(M-1)}$, aligns with that of the SlotContrast~\cite{slotcontrast} for a fair comparison.
We set the new hyperparameters consistently for all datasets: the repulsion coefficient $\beta=0.2$ and the SSIM3D loss coefficient $\lambda_{\text{SSIM3D}}=0.05$. 
We adopt the same coefficient $\lambda_{\text{SSC}}$ used in \cite{slotcontrast} for a fair comparison.
All metrics are computed over the full video sequences~(24 frames for the MOVi datasets and up to 76 frames for YouTube‑VIS).  
2 NVIDIA RTX A6000 GPUs were used for all experiments.
All benchmarking results are averaged over three runs.




\begin{table}[t]
\centering
\small
\vspace{-0.3cm}
\caption{
\textbf{Results on real-world YouTube-VIS dataset.}
}
\label{tab:ytvis}
\vspace{-0.15cm}
\renewcommand{\arraystretch}{0.9}
\setlength{\tabcolsep}{16.6pt}
\begin{tabular}{l cc}
\hlineB{2.5}
\multirow{2}{*}{\textbf{Method}} &
\multicolumn{2}{c}{\textbf{YouTube‑VIS}} \\ 
\cmidrule(lr){2-3}
& FG‑ARI\,$\uparrow$ & mBO\,$\uparrow$ \\ 
\hlineB{2.5}
STEVE~\cite{steve} & 15.0 & 19.1 \\
VideoSAUR~\cite{videosaur}  &  28.9 & 26.3 \\
VideoSAURv2~\cite{videosaur}  & 31.2 & 29.7 \\
SlotContrast~\cite{slotcontrast} & 38.0 & 33.7 \\
\rowcolor{slotrow}
SlotCurri~(\textbf{Ours}) & \textbf{44.8}$_{\pm{1.2}}$ & \textbf{35.5}$_{\pm{2.2}}$ \\
\hlineB{2.5}
\end{tabular}
\vspace{-0.2cm}
\end{table}

\subsection{Comparison with the State-of-the-art Methods}
In Tab.~\ref{tab:ytvis} and Tab.~\ref{tab:movi}, we benchmark SlotCurri against recent VOCL approaches~\cite{savi, steve, videosaur, slotcontrast}. 
SlotCurri delivers large improvements in both real‑world and synthetic datasets by addressing over-fragmentation in VOCL. 
Specifically, the mitigation of over-fragmentation is clearly validated by the FG-ARI metric; FG-ARI particularly penalizes fragmenting one GT object into multiple slots, so improvements indicate reduced over-fragmentation.
The substantial FG-ARI gains achieved by SlotCurri~(while maintaining SlotContrast's~\cite{slotcontrast} mBO on synthetic datasets and even improving on YouTube-VIS) therefore provide direct evidence that our method effectively reduces the targeted over-fragmentation.
We attribute this gain to our coarse‑to‑fine curriculum: by starting with very few slots, the model first learns broad semantic groupings, then allocates additional slots only where reconstruction error remains high. 
This ensures capacity is introduced exactly where it is most needed; slots that already model a region well keep their original assignments, preventing further fragmentation, whereas newly spawned slots focus on under-represented areas, refining coarse groupings into clear entity-level representations.

\begin{table}[t]
\vspace{-0.3cm}
\centering
\small
\caption{
\textbf{Results on synthetic MOVi-C and MOVi-E datasets.}
}
\label{tab:movi}
\vspace{-0.15cm}
\renewcommand{\arraystretch}{0.9}
\setlength{\tabcolsep}{3.2pt}
\begin{tabular}{l cc cc}
\hlineB{2.5}
\multirow{2}{*}{\textbf{Method}} &
\multicolumn{2}{c}{\textbf{MOVi‑C}} &
\multicolumn{2}{c}{\textbf{MOVi‑E}} \\
\cmidrule(lr){2-3}\cmidrule(lr){4-5}
& FG‑ARI\,$\uparrow$ & mBO\,$\uparrow$
& FG‑ARI\,$\uparrow$ & mBO\,$\uparrow$ \\ 
\hlineB{2.5}
SAVi~\cite{savi} & 22.2 & 13.6 & 42.8 & 16.0 \\
STEVE~\cite{steve} & 36.1 & 26.5 & 50.6 & 26.6 \\
VideoSAUR~\cite{videosaur} & 64.8$_{\pm1.2}$ & \textbf{38.9}$_{\pm0.6}$ & 73.9$_{\pm1.1}$ & \textbf{35.6}$_{\pm0.5}$ \\
VideoSAURv2~\cite{videosaur} & --   & --   & 77.1 & 34.4 \\
SlotContrast~\cite{slotcontrast} & 69.3 & 32.7 & 82.9 & 29.2 \\ 
\rowcolor{slotrow}
SlotCurri~(\textbf{Ours}) & \textbf{77.6}$_{\pm0.9}$ & 32.8$_{\pm0.2}$ & \textbf{83.7}$_{\pm0.2}$ & 28.9$_{\pm0.7}$   \\
\hlineB{2.5}
\end{tabular}
\vspace{-0.3cm}
\end{table}
Yet, the performance gain of SlotCurri on MOVi-E is relatively modest. 
We attribute this to a mismatch between our method's core design and the dataset's primary challenge.
SlotCurri is explicitly designed to resolve over-fragmentation, a common failure mode where large objects are incorrectly split into multiple slots. 
In contrast, our analysis suggests that a primary challenge in MOVi-E is under-fragmentation, which requires the model to perform fine-scale partitioning of many small, distinct objects. 
These results thus suggest that our SlotCurri has limitations in resolving under-fragmentation; extending our SlotCurri to address under-fragmentation remains an important area for future work.
Qualitative discussion is in the Appendix.


Furthermore, we observe that VideoSAUR~\cite{videosaur} attains higher mBO on synthetic datasets.
We attribute this to its direct modeling of motion patterns, which may be particularly advantageous given the relatively limited, near-rigid transformations in synthetic settings. 
In contrast, on the more variable, real-world YouTube-VIS~(objects exhibit higher degrees of freedom), SlotCurri achieves state-of-the-art FG-ARI and mBO. 
This contrast highlights that our approach is especially effective under realistic, challenging conditions.

\setlength{\columnsep}{4pt} 
\begin{wraptable}{h}{0.20\textwidth}
\centering
\small
\vspace{-0.53cm}
\caption{
\textbf{Image FG-ARI compared against baselines targeting over-fragmentation} on MOVi-C and MOVi-E datasets.
}
\label{tab:image}
\vspace{-0.2cm}
\renewcommand{\arraystretch}{0.9}
\setlength{\tabcolsep}{0.9pt}
\begin{tabular}{l c c}
\hlineB{2.5}
\multirow{2}{*}{\textbf{Method}} &
\multicolumn{1}{c}{\textbf{C}} &
\multicolumn{1}{c}{\textbf{E}} \\
\cmidrule(lr){2-3}
& \multicolumn{2}{c}{Image FG‑ARI} \\
\hlineB{2.5}
AdaSlot~\cite{adaptiveslot} & 75.6 & 76.7 \\
SOLV~\cite{aydemir2023self} & - & 80.8  \\
\rowcolor{slotrow}
SlotCurri & \textbf{81.6} & \textbf{84.9} \\
\hlineB{2.5}
\end{tabular}
\vspace{-0.3cm}
\end{wraptable}
\setlength{\columnsep}{10pt} 
Finally, in Tab.~\ref{tab:image}, we compare SlotCurri with methods that aim to address object over-fragmentation. 
Using per-frame metrics with their reported scores, our results show a notable performance gap. 
This demonstrates SlotCurri's advantage in preventing object over-fragmentation from the outset.




\begin{table}[t]
\centering
\small
\vspace{-0.2cm}
\caption{
    \textbf{Ablation study on YouTube-VIS.}
    Our baseline is SlotContrast, reproduced using official code.
    R-G denotes reconstruction-guided slot curriculum, Struct is structure-aware reconstruction loss, and Cycle indicates the cyclic inference. 
}
\label{tab:ablation}
\vspace{-0.12cm}
\renewcommand{\arraystretch}{0.89}
\setlength{\tabcolsep}{7.2pt}
\begin{tabular}{c c c c cc}
\hlineB{2.5}
\multicolumn{2}{c}{Curriculum} & \multirow{2}{*}{Struct.} & \multirow{2}{*}{Cycle} & \multicolumn{2}{c}{YouTube‑VIS} \\ 
\cmidrule(lr){0-1}\cmidrule(lr){5-6}
Simple & R-G & & & FG‑ARI\,$\uparrow$ & mBO\,$\uparrow$ \\ 
\hlineB{2.5}
- & - & - & - & 36.1 & 32.7 \\
\cmark & - & - & - & 38.8 & 32.3 \\
\cmark & - & - & \cmark & 40.3	& 32.3 \\
- & \cmark & - & - & 42.6 & 33.7 \\
- & \cmark & \cmark & - & 43.6 & 35.2 \\
- & \cmark & \cmark & \cmark & 44.8 & 35.5 \\
\hlineB{2.5}
\end{tabular}
\vspace{-0.15cm}
\end{table}

\subsection{Further Studies}
All studies are conducted on the YouTube-VIS dataset.

\textbf{Component Ablation.}
We conduct an ablation study summarized in Tab.~\ref{tab:ablation}.
Our baseline is the reproduced version of SlotContrast.
First, we evaluate the impact of curriculum learning in the context of VOCL, where object cues are absent and slots tend to over-fragment objects.
As shown in the 2nd and 4th rows, both simple curriculum learning and our reconstruction-guided~(R-G) curriculum learning lead to substantial improvements over the baseline, with our R-G curriculum achieving gains of up to 6.5 points in FG-ARI.
This significant gain stems from two key factors:
(1) starting with a smaller number of slots encourages contextually meaningful grouping early on, which anchors representations around optimal slot counts and
(2) as capacity grows, additional slots are allocated to regions requiring finer granularity, enabling more effective semantic slot separation.
Also, SSIM loss further enhances performance since well-separated semantics in parent slots are crucial for ensuring that, after expansion, child slots inherit clear semantic boundaries.
Finally, the cyclic inference improves the grouping of coherent semantics across slots by incorporating long-range contextual cues from both temporal directions, with negligible inference overhead.
We qualitatively illustrate the impact of each component in the Appendix.


\begin{table}[!t]
  \centering
  \footnotesize
  \setlength{\tabcolsep}{2.5pt}
\renewcommand{\arraystretch}{0.92}
  \caption{
  \textbf{Hyperparameter sensitivity analysis on YouTube-VIS.}
  Bolded parameters denote default configurations.}
  \label{Tab.ablation_curriculum}
  \vspace{-0.3cm}
  \begin{subtable}[t]{0.28\linewidth}
    \centering
    \caption{Impact of the number of curriculum stages~($M$).}
    \label{Tab.curriculum_M}
    \begin{tabular}{l cc}
    \hlineB{2.5}
    \multirow{2}{*}{$M$} &
    \multicolumn{2}{c}{\textbf{YouTube‑VIS}} \\ 
    \cmidrule(lr){2-3}
    & FG‑ARI & mBO \\ 
    \hlineB{2.5}
    2 & 41.5 & 36.3 \\
    \textbf{3} & 44.8 & 35.5 \\
    4 & 44.7 & 34.2 \\
    \hlineB{2.5}
    \end{tabular}
  \end{subtable}\hfill
  \begin{subtable}[t]{0.32\linewidth}
    \centering
    \caption{Impact of the repulsion coefficient~($\beta$) for spawning child slots.}
    \label{Tab.curriculum_beta}
    \begin{tabular}{l cc}
    \hlineB{2.5}
    \multirow{2}{*}{$\beta$} &
    \multicolumn{2}{c}{\textbf{YouTube‑VIS}} \\ 
    \cmidrule(lr){2-3}
    & FG‑ARI & mBO \\  
    \hlineB{2.5}
    0.1 & 42.8 & 33.9 \\
    \textbf{0.2} & 44.8 & 35.5\\
    0.3 & 40.2 & 33.2 \\
    \hlineB{2.5}
    \end{tabular}
  \end{subtable}\hfill
    \begin{subtable}[t]{0.36\linewidth}
    \centering
    \caption{Impact of the magnitude of SSIM loss~($\lambda_{\text{SSIM3D}}$).}
    \label{Tab.curriculum_ssim}
    \begin{tabular}{l cc}
    \hlineB{2.5}
    \multirow{2}{*}{$\lambda_{\text{SSIM3D}}$} &
    \multicolumn{2}{c}{\textbf{YouTube‑VIS}} \\ 
    \cmidrule(lr){2-3}
    & FG‑ARI & mBO \\ 
    \hlineB{2.5}
    0.02 & 44.2 & 34.2 \\
    \textbf{0.05} & 44.8 & 35.5 \\
    0.07 & 43.9 & 35.0 \\
    \hlineB{2.5}
    \end{tabular}
  \end{subtable}\hfill
  \vspace{-0.4cm}
\end{table}

\begin{figure*}[t]
\centering
\vspace{-0.2cm}
\includegraphics[width=0.90\textwidth]{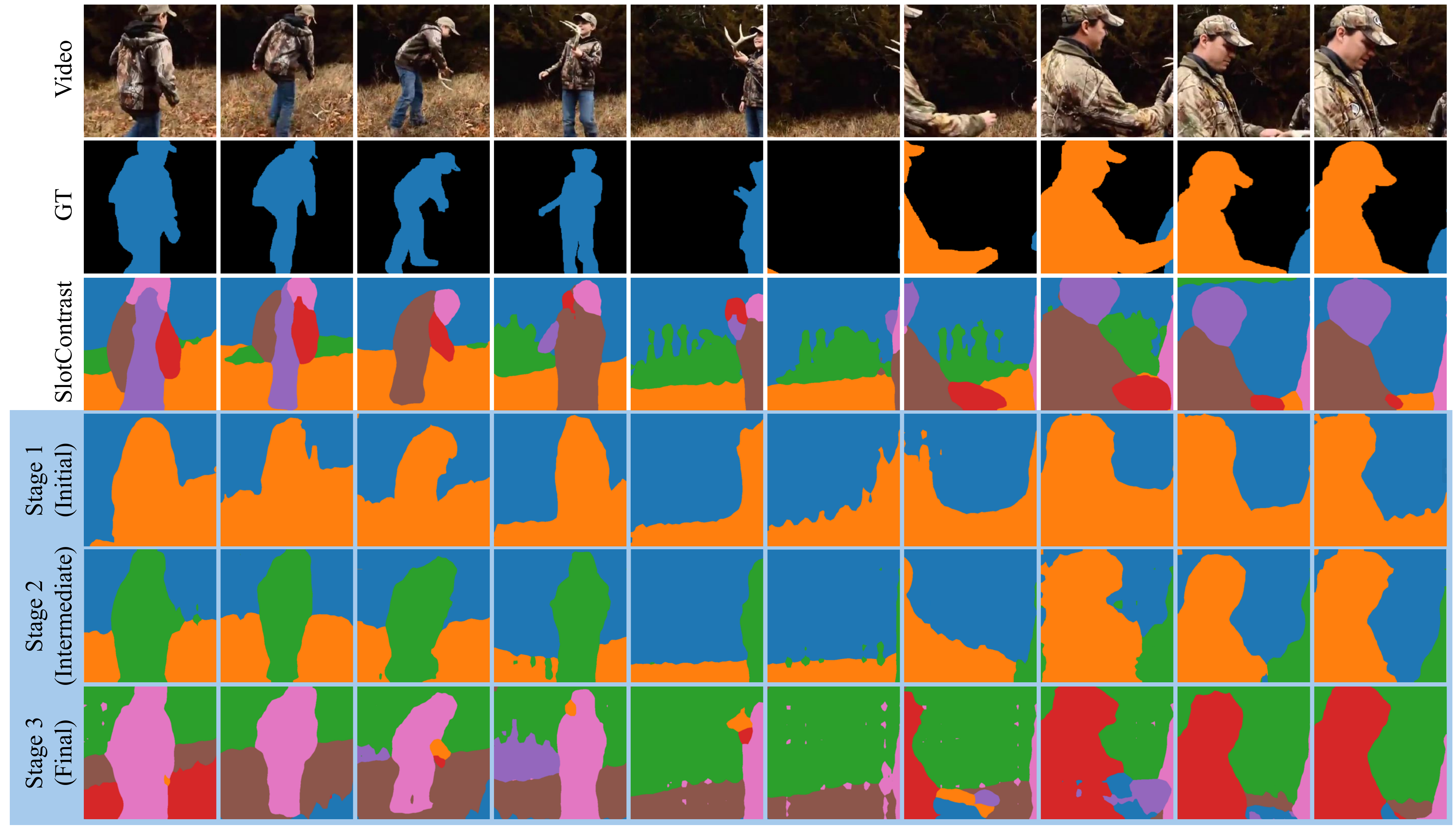} 
\vspace{-0.1cm}
\caption{
\textbf{Qualitative results on YouTube-VIS dataset.}
We present visualizations of video frames along with their corresponding GT masks and the predicted masks from SlotContrast and our method.
The examples in the fourth to sixth rows (highlighted with a blue background) illustrate how our slot representations evolve across different stages of the curriculum.
}
\label{Fig.qualitative}
\vspace{-0.25cm}
\end{figure*}

\textbf{Varying the Number of Curriculum Stages $M$.} 
As shown in Tab.~\ref{Tab.ablation_curriculum}~(a), setting $M$=3 yields the highest FG-ARI score of 44.8. 
A single stage transition~($M$=2) is insufficient for the model to learn coarse-to-fine representations, resulting in a lower FG-ARI of 41.5. 
Conversely, increasing the number of stages to $M$=4 leads to a performance drop, suggesting that three stages are sufficient to learn an effective coarse-to-fine representation.
Also, while $M$=2 achieves the highest mBO, our primary goal is to mitigate over-fragmentation, for which FG-ARI is a more direct measure. 
Therefore, we select $M$=3 as the default setting.

\textbf{Varying the Slot Repulsion Coefficient $\beta$.} 
The parameter $\beta$ scales the magnitude of Gaussian noise applied to a parent slot embedding when spawning a new child slot. 
This perturbation encourages the new slot to explore under-represented regions of the feature space rather than simply duplicating its parent's representation. 
The results in Tab.~\ref{Tab.ablation_curriculum}~(b) show a clear trend. 
A small $\beta$ is suboptimal as the new slot may be too similar to its parent to effectively capture novel features, whereas a large $\beta$ also degrades performance, likely because excessive noise disrupts the transfer of useful information from the parent slot, making the initialization too random. 
We find that $\beta$=0.2 provides the best trade-off, achieving the highest scores for both FG-ARI and mBO, and thus we use this value in our final model.

\textbf{Varying the Coefficient of Structure-Aware Reconstruction loss $\lambda_{\text{SSIM3D}}$.}
In Tab.~\ref{Tab.ablation_curriculum}~(c), we analyze the impact of $\lambda_{\text{SSIM3D}}$. 
We find that incorporating the SSIM3D loss~(\textit{i.e.}, $\lambda_{\text{SSIM3D}}>$\;0) is beneficial, as it encourages the model to learn sharper and more accurate semantic boundaries. 
However, we observe a clear performance degradation when this coefficient is set too high. 
This suggests that while enforcing structural awareness is advantageous, an excessive focus on boundary-level fidelity can detract from the primary task, leading to a drop in overall performance.

\begin{table}
\vspace{-0.2cm}
\small
\caption{
    \textbf{Analysis of object discovery quality against SlotContrast~(SC).}
}
\footnotesize
\renewcommand{\arraystretch}{1}
\setlength{\tabcolsep}{1.7pt}
\vspace{-0.2cm}
\begin{subtable}[t]{0.19\textwidth}
\caption{Object Identification Recall~(OIR@$\rho$). 
Percentage of GT objects for which there exists at least one slot with IoU $\ge \rho$ (higher is better).}
\label{tab:oir}
\begin{tabular}{lccc}
\hlineB{2.5}
Method & @0.3 & @0.5 & @0.7 \\
\hlineB{2.5}
SC~\cite{slotcontrast} & 48.0 & 24.9 & 4.6 \\
Ours         & \textbf{52.2} & \textbf{30.3} & \textbf{7.3} \\
\hlineB{2.5}
\end{tabular}
\end{subtable}
\hfill
\begin{subtable}[t]{0.275\textwidth}
\vspace{-0.4cm}
\centering
\footnotesize
\renewcommand{\arraystretch}{0.85}
\setlength{\tabcolsep}{5.3pt}
\caption{Average number of slots per GT object~(Degree of over-fragmentation).
Each slot is considered to represent a GT object only if at least an $\rho$ fraction of its area lies within that object.
Computed only over objects with $\ge$ 1 representing slot, and higher values indicate more fragmentation. 
}
\label{tab:overfrag}
\begin{tabular}{lccc}
\hlineB{2.5}
Method & @0.3 & @0.5 & @0.7 \\
\hlineB{2.5}
SC~\cite{slotcontrast} & 1.38 & 1.38 & 1.46 \\
Ours & \textbf{1.29} & \textbf{1.26} & \textbf{1.36} \\
\hlineB{2.5}
\end{tabular}
\end{subtable}
\vspace{-0.4cm}
\end{table}



\textbf{Analysis on Object Identification of SlotCurri.}
In Tab.~\ref{tab:oir}, we measure how reliably the slot representation identifies ground-truth~(GT) instances, with a newly introduced Object Identification Recall~(OIR) metric.
Concisely, OIR@$\rho$ is the fraction of foreground GT instances for which at least one slot attains IoU$\ge \rho$ with the instance mask (see Appendix for details); higher is better, and it isolates object coverage irrespective of fragmentation.
As shown, our approach improves OIR by $+4.2/+5.4/+2.7$ points at $\rho=0.3/0.5/0.7$, respectively, indicating a markedly higher likelihood that at least one slot correctly captures each GT object under the same IoU thresholds.


\textbf{Over-fragmentation.}
In addition, we evaluate object over-fragmentation, the main challenge this paper targets. 
From Tab.~\ref{tab:oir}, we observe that even the state-of-the-art methods may occasionally miss an object altogether in challenging real-world videos.
To assess over-fragmentation fairly, we report the degree of over-fragmentation~(DOF) only on GT objects that are successfully detected by at least one slot.
DOF is defined as the average number of slots per detected object.
The lower values~(closer to 1) indicate fewer unnecessary splits of the same object.
As shown in Tab.~\ref{tab:overfrag}, our method reduces over-fragmentation from 1.38 to 1.26 slots per object when $\rho=0.5$, demonstrating that the same object is less likely to be split across multiple slots.

\subsection{Qualitative Results}
Fig.~\ref{Fig.qualitative} provides a qualitative overview of the representations learned by our SlotCurri.
Each row presents the original RGB frames, GT masks, and predictions from SlotContrast~\cite{slotcontrast}, followed by visualizations of slot masks at three different stages of SlotCurri.
As shown in the third row, SlotContrast tends to over-fragment objects~(\textit{e.g.}, a person is split into torso, head, and arm) since SlotContrast declares all slots from the beginning.
Also, we find that it struggles to identify different identities of similar appearance~(\textit{e.g.}, different individuals in the 5th and 7th frames are matched to the same slot across frames).
In contrast, the final row shows that our SlotCurri effectively alleviates over-fragmentation by learning identity-specific representations. 
It also separates visually similar identities more effectively by allocating additional capacity to slots that exhibit high reconstruction error due to their broad coverage of different entities.

In addition, we illustrate the progression across curriculum stages~(the 4th–6th rows).
Specifically, training begins with only two slots in the 4th row, where the model captures coarse regions that often span multiple contexts. 
As training progresses and new slots are introduced, these initially broad and heterogeneous regions are recursively partitioned into finer, entity-level segments.


\section{Conclusion}
We propose SlotCurri to address over-fragmentation in VOCL. 
Training starts with a minimal slot budget, expands capacity only where reconstruction error remains high, and initializes each newborn slot with distance‑aware noise so that it explores under‑represented regions instead of redundantly splitting well‑modeled ones. 
An SSIM-driven structural loss assists in enabling more precise slot expansion by sharpening semantic boundaries in each slot.
Lastly, our cyclic inference strengthens early-frame representations by aggregating contextual cues with negligible overhead and no additional training.
Taken together, our results establish SlotCurri as a practical training paradigm for VOCL and an effective solution to over-fragmentation.
Future work will investigate scene‑adaptive slot schedules and multi‑scale slot hierarchies to advance video object discovery further.

\noindent\textbf{Broader Impact.}
This work establishes compact and consistent slots by mitigating over-fragmentation.
We expect this to benefit downstream tasks by enabling accurate and efficient object-level dynamics modeling, relationship reasoning, and video editing.

\noindent\textbf{Acknowledgements}
This work was supported in part by MSIT/IITP (No. RS-2022-II220680, RS-2020-II201821, RS-2019-II190421, RS-2024-00459618, RS-2024-00360227, RS-2024-00437633, RS-2024-00437102, RS-2025-25442569), MSIT/NRF (No. RS-2024-00357729), and KNPA/KIPoT (No. RS-2025-25393280).

\maketitlesupplementary
\renewcommand{\thesection}{A}   
\renewcommand{\thetable}{A\arabic{table}}   
\renewcommand{\thefigure}{A\arabic{figure}}
\setcounter{section}{0}
\setcounter{table}{0}
\setcounter{figure}{0}


\renewcommand{\thesection}{A}   
\section{Datasets}
To benchmark our method against existing video object-centric learning approaches, we conducted evaluations on three widely used datasets: YouTube-VIS 2021, MOVi-C, and MOVi-E.
To ensure a fair comparison, we adopted the identical data splits and preprocessing methodology as SlotContrast~\cite{slotcontrast}.
The YouTube-VIS 2021 dataset~\cite{ytvis} is a large-scale, real-world video instance segmentation benchmark derived from the YouTube-VOS dataset~\cite{youtubevos}. 
Following our baseline~\cite{slotcontrast}, we utilized 2,775 training videos and 210 validation videos at an image resolution of $518 \times 518$ pixels.
MOVi-C and MOVi-E are synthetic datasets generated using Kubric~\cite{greff2022kubric}. 
MOVi-C comprises scenes with up to 10 scanned 3D objects, realistic backgrounds, and free camera movement. 
MOVi-E increases the complexity significantly, featuring up to 23 objects per scene and randomized camera trajectories, making it particularly challenging for unsupervised object discovery. 
Both MOVi datasets include 9,750 training videos and 250 validation videos each, rendered at a resolution of $336 \times 336$ pixels.
All experiments were conducted using two NVIDIA RTX A6000 GPUs and an Intel Xeon Gold 5220R CPU.

\renewcommand{\thesection}{B}   
\section{Object Identification Recall}
In our manuscript, we introduced Object Identification Recall~(OIR) to measure how reliably the learned slots identify ground-truth~(GT) instances.
Specifically, OIR@$\rho$ represents the fraction of foreground GT instances for which at least one slot attains IoU$\ge \rho$ with the instance mask.
Let $g$ denote the GT mask of an object at a specific frame, and let $\{m_k\}_{k=1}^S$ denote the predicted masks from all $S$ slots, we compute $\mathrm{IoU}(g,m_k)=\frac{|g \cap m_k|}{|g \cup m_k|}$ and report the OIR at threshold $\rho$~(OIR@\,$\rho$):
\begin{equation}
\text{OIR@}\rho \!=\!
\!\frac{1}{\left|\mathcal{G}_{\text{valid}}\right|}
\sum_{g \in \mathcal{G}_{\text{valid}}}\!
\mathbf{1}\!\left[ \max_{k} \, \mathrm{IoU}\!\left(g,m_k\right) \ge \rho \right],
\label{eq:oir}
\end{equation}
where $\mathcal{G}_{\text{valid}}$ is the set of GT objects with non-zero area~(excluding the background class).
OIR measures whether each foreground object is covered by at least one slot or not.

\renewcommand{\thesection}{C}  
\section{Degree of Over-Fragmentation}
To precisely assess the over-fragmentation, we introduced the Degree of Over-Fragmentation~(DOF).
In this section, we illustrate DOF in detail.
We say that the $k$-th slot represents a specific GT object if at least an $\rho$ fraction of the slot area~($m_k$) lies within the GT mask $g$:
\begin{equation}
\frac{|m_k \cap g|}{|m_k|} \;\ge\; \rho 
\;\;\Longrightarrow\;\; m_k \rightarrow g .
\label{eq:assign}
\end{equation}
Let $\mathcal{G}_{\mathrm{det}}=\{\, g \in \mathcal{G}_{\text{valid}} \;|\; \exists k:\, m_k \rightarrow g \,\}$ denote the set of detected objects.
We report the average number of slots that are assigned to each detected object:
\begin{equation}
\text{DOF}
\;=\;
\frac{1}{|\mathcal{G}_{\mathrm{det}}|}
\sum_{g \in \mathcal{G}_{\mathrm{det}}}
\Big|\,\{\,k:\, m_k \rightarrow g\,\}\,\Big| ,
\label{eq:frag}
\end{equation}
where lower values~(closer to 1) indicate fewer unnecessary splits of the same object.
Note that the size of $|\mathcal{G}_{\mathrm{det}}|$ can be different depending on the number of identified GT objects.
Also, we only evaluate the DOF on identified GT objects since this isolates the over-fragmentation analysis from object discovery failures; if we include unidentified objects, which have a slot count of zero, it would artificially deflate the average and confound the metric with detection recall~(which is already measured by OIR).

\begin{table}[h]
\centering
\small
\caption{\textbf{Average number of GT objects per slot~(Degree of under-fragmentation).} 
A GT object is matched to a slot if $\text{IoU} \ge \rho$. 
Higher values indicate that single slots contain multiple objects~(severe under-fragmentation).
}
\label{tab:underfrag}
\renewcommand{\arraystretch}{1}
\setlength{\tabcolsep}{10pt}
\begin{tabular}{l ccc}
\toprule
Method & @0.3 & @0.5 & @0.7 \\
\midrule
SContrast & 1.23 & 1.21 & 1.19 \\
\rowcolor{gray!20}
Ours & \textbf{1.22} & \textbf{1.18} & \textbf{1.15} \\
\bottomrule
\end{tabular}
\end{table}
\renewcommand{\thesection}{D}  
\section{Degree of Under-Fragmentation}
\label{sec:duf}
Complementary to over-fragmentation, we strictly assess the under-segmentation issue using the Degree of Under-Fragmentation~(DUF).
While over-fragmentation splits one object into many slots, under-fragmentation occurs when a single slot erroneously merges multiple distinct objects.
Analogous to Eq.~\ref{eq:assign}, we consider a slot to capture a ground-truth object if their Intersection-over-Union~(IoU) exceeds a specified threshold $\rho$.
DUF is then defined as the average count of ground-truth objects assigned to each matched slot.
In this metric, a value of 1.0 represents the ideal one-to-one correspondence, while higher values indicate a failure to disentangle distinct instances~(i.e., severe under-fragmentation).

The quantitative results are presented in Tab.~\ref{tab:underfrag}.
Notably, SlotCurri achieves the lower DUF scores across all thresholds compared to the baseline.
This indicates that our SlotCurri does not merely shift the error distribution from over- to under-fragmentation but fundamentally improves the distinctness and quality of object disentanglement.

\begin{figure}[t!]
\centering
\includegraphics[width=1.\columnwidth]{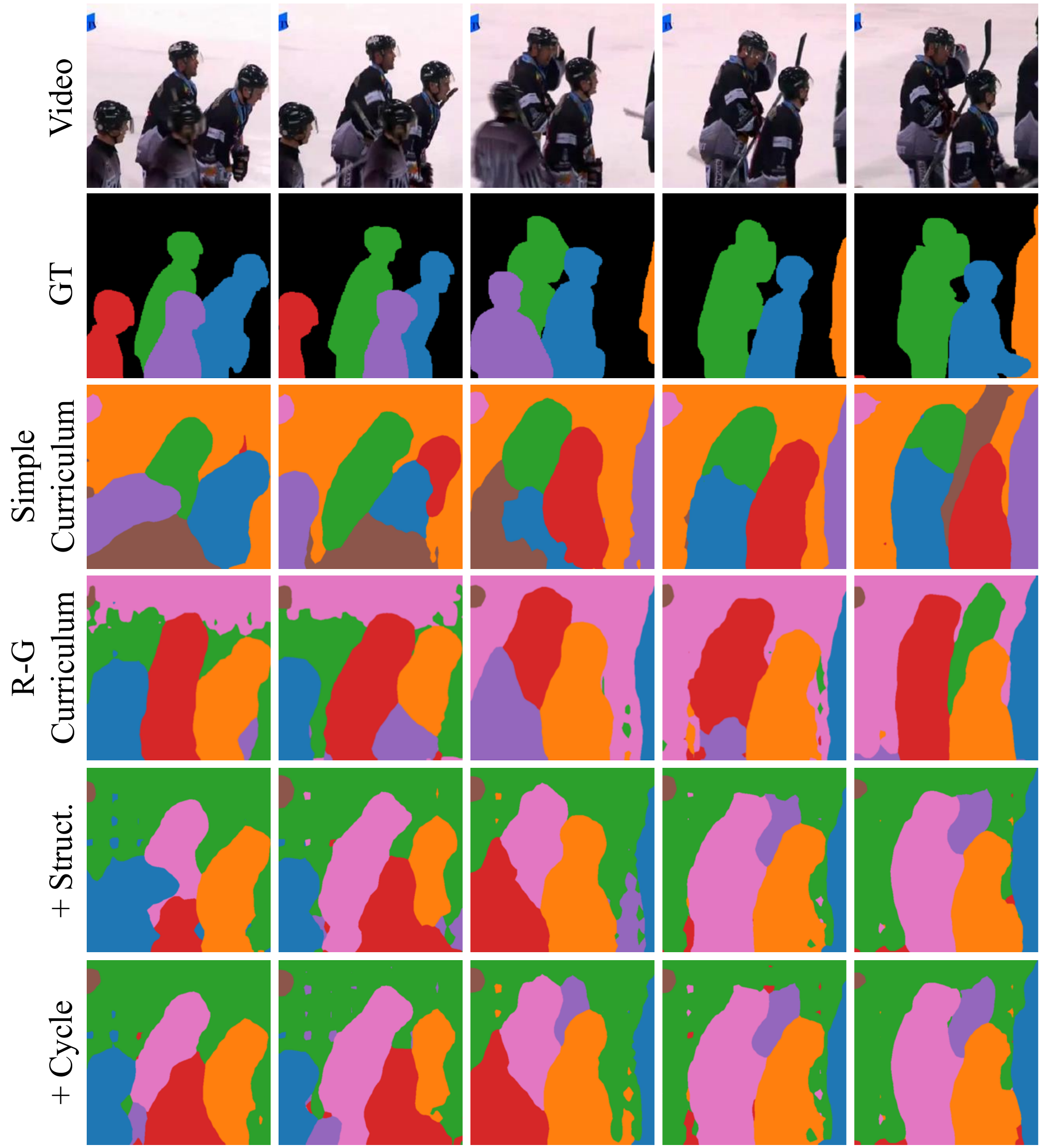} 
\caption{
\textbf{Qualitative analysis of the impact of each component.}
}
\label{Fig.viz_ablation}
\end{figure}

\renewcommand{\thesection}{E}   
\section{Qualitative Ablation Study}
This section qualitatively illustrates the impact of each component with the visualizations in Fig.~\ref{Fig.viz_ablation}.
The R-G curriculum mitigates over-fragmentation~(4th row), while the SSIM objective sharpens object boundaries~(\textit{e.g.}, pink, red, and orange masks in 2nd column of 5th row) and leverages compositional structures to clearly separate distinct entities~(\textit{e.g.}, pink and red masks in 1st column of 5th row).
Furthermore, cyclic inference effectively incorporates long-range contextual cues, enhancing slot compactness even in early frames~(\textit{e.g.}, blue and red masks in 1st column of last row).

\renewcommand{\thesection}{F} 
\section{Analysis of the Accelerated Slot Schedule}
\begin{table}[h]
\centering
\small
\caption{\textbf{Ablation study on the scheduling strategies on YouTube-VIS.}
}
\label{tab:schedule_ablation}
\renewcommand{\arraystretch}{1}
\setlength{\tabcolsep}{19pt}
\begin{tabular}{l cc}
\toprule
\multirow{2}{*}{\textbf{Slot Scheduling}} &
\multicolumn{2}{c}{\textbf{YouTube‑VIS}} \\ 
\cmidrule(lr){2-3}
& FG‑ARI\,$\uparrow$ & mBO\,$\uparrow$ \\ 
\midrule
Decelerated & 38.2 & 32.9 \\
Linear & 43.4 & 34.2 \\
\rowcolor{gray!20}
Accelerated & 44.8 & 35.5 \\
\bottomrule
\end{tabular}
\end{table}
Our SlotCurri framework, as defined in Eq.~(2), employs an accelerated slot schedule where the number of new slots added increases quadratically at each curriculum stage. This design is based on the hypothesis that the learning process benefits from allocating progressively more representational capacity to later, more complex refinement stages.
To validate this hypothesis, we conduct an ablation study comparing our accelerated schedule against two alternatives, all starting with $K_\text{init} = 2$ and ending with the same final slot count~($K_\text{final}$): (1) Linear Schedule and (2) Decelerated Schedule.

To illustrate, for the linear schedule, the total number of new slots is distributed as evenly across the stages, while the decelerated schedule adds the majority of slots in the first expansion stage and progressively fewer in later stages (the inverse of our accelerated approach).
As shown in Tab.~\ref{tab:schedule_ablation}, the choice of schedule is critical. 
The decelerated schedule, which introduces high capacity too early, yields the smallest gain. 
This suggests that adding fine-grained capacity before coarse-level semantics have stabilized is detrimental, leading to premature over-fragmentation. 
The linear schedule performs better, but it is significantly outperformed by our accelerated approach.

This result provides strong evidence for our coarse-to-fine learning hypothesis, which involves two distinct phases. 
The first stage involves learning to anchor broad, stable, and semantically coherent regions, which requires relatively low slot capacity. 
Then, the second stage is to partition the fine-grained details within these coarse regions, which is a combinatorially more complex task, requiring a much larger representational budget to capture the explosion of new, smaller entities.
Consequently, we observe that the accelerated schedule is effective as it mirrors this learning dynamic.

\renewcommand{\thesection}{G}   
\section{Analysis of the Reconstruction-Guided Slot Spawning Criterion}
\begin{table}[h]
\centering
\small
\caption{
\textbf{Analysis of slot spawning criteria on YouTube-VIS.}
We compare our proposed criterion~(Total Error Mass) against a scale-invariant alternative~(Area-Normalized Error). 
}
\label{tab:splitcriteria}
\renewcommand{\arraystretch}{1}
\setlength{\tabcolsep}{14.7pt}
\begin{tabular}{l cc}
\toprule
\multirow{2}{*}{\textbf{Slot Splitting Criteria }} &
\multicolumn{2}{c}{\textbf{YouTube‑VIS}} \\ 
\cmidrule(lr){2-3}
& FG‑ARI\,$\uparrow$ & mBO\,$\uparrow$ \\ 
\midrule
Area Norm & 40.4 & 32.7 \\
\rowcolor{gray!20}
Total Error & 44.8 & 35.5 \\
\bottomrule
\end{tabular}
\end{table}

\begin{figure}[h]
\centering
\includegraphics[width=1.\columnwidth]{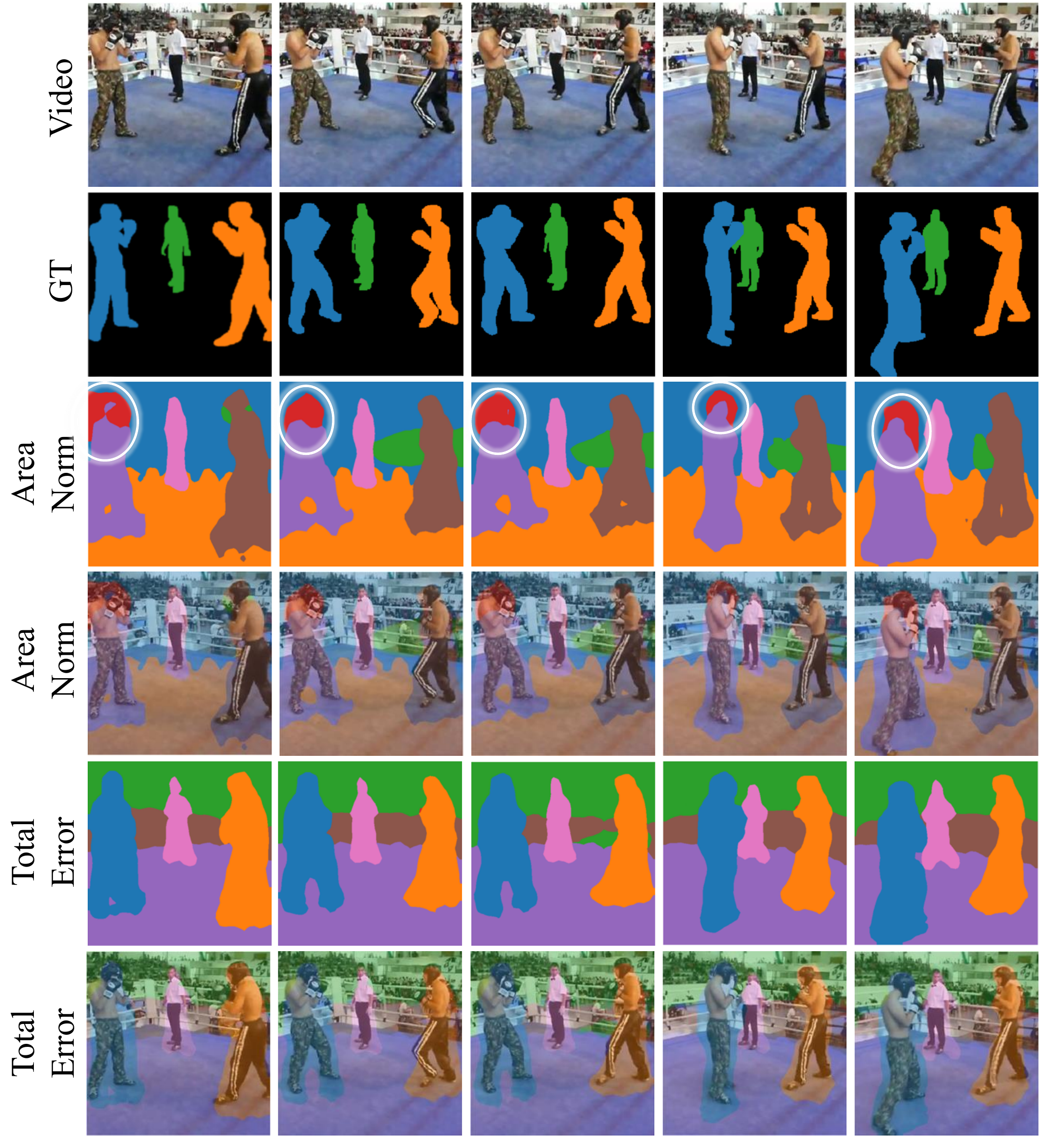} 
\caption{
    \textbf{Qualitative Comparison of Spawning Criteria.}
}
\label{Fig.viz_ablation_spawning}
\end{figure}

Our reconstruction-guided curriculum allocates new slots by duplicating parent slots responsible for the highest reconstruction error, $\delta^{(k)}$.
This criterion measures the total error mass accumulated by a slot, weighted by its slot's reconstruction weights in the decoder~($\alpha^{(k,t,h,w)}$) across all spatio-temporal positions.

The primary advantage of this formulation is that it provides a usage-aware capacity allocation; slots that are idle, weakly used, or already perfectly reconstructed will have a negligible $\delta^{(k)}$ and are thus ineligible for splitting. 
This mechanism is crucial as it prevents the model from wasting capacity by unnecessarily fragmenting auxiliary or well-modeled regions, focusing refinement only on slots that are actively contributing to the reconstruction.

An alternative is to form a scale-invariant criterion by normalizing $\delta^{(k)}$ with the slot's total ㅈeight mass~(\textit{i.e.}, area normalization). 
While this could, in theory, find high-error slots regardless of their size, it introduces a critical instability. 
Since the reconstruction weights~($\alpha$) are continuous, an idle or weakly used slot with a near-zero mass can yield an arbitrarily high normalized error from minor noise. 
To validate this, we empirically tested this area-normalized variant in Tab.~\ref{tab:splitcriteria}, which shows that this variant markedly underperformed on the video benchmark.
Also, Fig.~\ref{Fig.viz_ablation_spawning} illustrates a common failure mode of this area-normalized variant. 
We observe that semantically distinct regions, such as the head and torso of a person, are often split into different slots. 
This suggests that normalizing by area can over-emphasize local error in semantically distinct sub-regions~(like the head), even when they belong to the same object, leading to undesirable fragmentation.

Our findings suggest that, in our VOCL setup, allocating capacity in proportion to the total error mass is a more robust and effective strategy by naturally balancing the model's focus between reconstruction fidelity and semantic importance. 
We acknowledge that for domains dominated by small, dense objects~(\textit{e.g.,} MOVi-E), hybrid criteria may offer further robustness, which we leave as a promising direction for future work.

\renewcommand{\thesection}{H}
\section{Chunk-wise Short Cyclic Inference}
\begin{wraptable}{h!}{0.24\textwidth}
\vspace{-0.4cm}
\centering
\small
\caption{
\textbf{Performance of chunk-wise short cyclic inference on YouTube-VIS.}
We compare the full cycle, no cycle~(\xmark), and short cycles constrained to each video chunk of $C$ frames~($C\in\{2, 3\}$) on YouTube-VIS.
}
\label{tab:cycle}
\renewcommand{\arraystretch}{1}
\setlength{\tabcolsep}{2.2pt}
\begin{tabular}{c cc}
\toprule
Cycle &
\multicolumn{2}{c}{\textbf{YouTube‑VIS}} \\ 
\cmidrule(lr){2-3}
\#Frames~($C$)& FG‑ARI\,$\uparrow$ & mBO\,$\uparrow$ \\ 
\midrule
\xmark & 43.6 & 35.2 \\
2 & 43.8 & 35.2 \\
3 & 43.9 & 35.1 \\
\rowcolor{gray!20}
full & 44.8 & 35.5 \\
\bottomrule
\end{tabular}
\vspace{-0.3cm}
\end{wraptable}
In the paper, we indicated that our cyclic inference is substantially more lightweight than the heavy encoder or decoder, leading to a negligible increase in inference time. 
This section further investigates the applicability of our method in streaming video applications where the latency is critical.
For long videos, applying the full cycle across all frames introduces a significant computational bottleneck, making it impractical for real-time streaming.

Therefore, we explore whether a short cycle~(where the signal propagates only to the next frame or a few subsequent frames before returning) can effectively replace the full cycle.
To this end, we experiment by processing the video in chunks of $C$ frames~(where $C$ is the chunk size) and constraining the cycle to operate only within these chunks. 
Specifically, we partition the video into sequential chunks, each containing a maximum of $C$ frames. 
The cyclic inference is then performed locally within each chunk. 
As the model processes frames and reaches the end of a chunk, a backward cyclic pass is initiated, propagating information back to the start of that same chunk.
This chunk-wise mechanism ensures that the cyclic inference is confined to a short, fixed-duration segment. 
Consequently, this approach is highly applicable to streaming scenarios, as it (1) strictly bounds the computational latency of the cyclic operation and (2) generates object slots on a per-chunk basis, removing the need to buffer or process the entire video before producing an output.
As shown in Tab.~\ref{tab:cycle}, while this chunk-wise cycle does not achieve the full performance of a complete cycle, it still demonstrates a clear improvement over the model with no cycle. 
This finding indicates that even a localized cyclic mechanism offers a practical trade-off, enhancing performance while maintaining viability for streaming applications.

\begin{figure*}[t]
\centering
\includegraphics[width=1\textwidth]{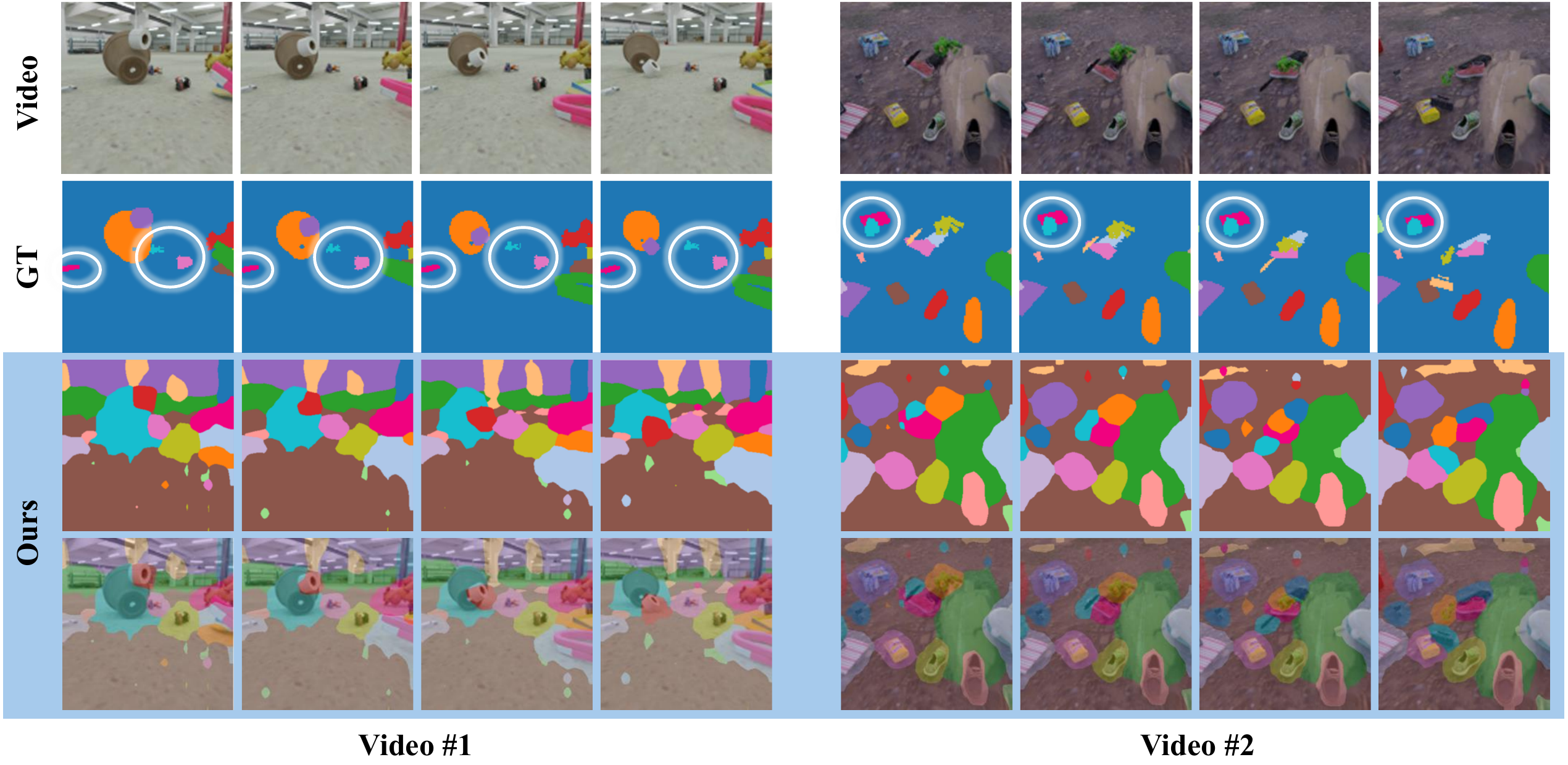} 
\caption{
\textbf{Qualitative results on MOVi-E dataset.}
We present visualizations of video frames along with their corresponding GT masks and prediction masks from our SlotCurri.
The white circles in the first row highlight the limitation of our work.
}
\label{Fig.movie}
\end{figure*}

\renewcommand{\thesection}{I} 
\section{Necessity of Structure Loss in Slot Curriculum Learning}
\begin{table}[h]
\centering
\small
\caption{\textbf{Effectiveness of structure-aware reconstruction loss}~($\mathcal{L}_{\text{SSIM3D}}$). 
$\ast$ represents the reproduced results.
}
\label{tab:ssim}
\renewcommand{\arraystretch}{1}
\setlength{\tabcolsep}{10.6pt}
\begin{tabular}{lc cc}
\toprule
\multirow{2}{*}{\textbf{Method}} & \multirow{2}{*}{$\mathcal{L}_{\text{SSIM3D}}$} &
\multicolumn{2}{c}{\textbf{YouTube‑VIS}} \\ 
\cmidrule(lr){3-4}
& & FG‑ARI\,$\uparrow$ & mBO\,$\uparrow$ \\ 
\midrule
SlotContrast$^\ast$ & - & 36.1 & 32.7 \\
SlotContrast & \checkmark & 36.4 & 32.7 \\
R-G Curriculum & - & 42.9 & 33.8 \\
\rowcolor{gray!20}
R-G Curriculum & \checkmark & 44.8 & 35.5 \\
\bottomrule
\end{tabular}
\end{table}
As discussed in the main manuscript, the commonly used mean-squared-error~(MSE) reconstruction loss tends to blur spatial details and obscure the true boundaries between objects. 
This issue becomes more problematic under our curriculum learning framework, as blurred boundaries lead to ambiguous slot representations that are further subdivided in subsequent stages. 
To validate this effect, we investigate the impact of incorporating our structure-aware reconstruction loss into the baseline SlotContrast~\cite{slotcontrast}. 
As shown in Tab.~\ref{tab:ssim}, our reconstruction-guided curriculum learning exhibits substantial gains of +2.3 points in FG-ARI and +1.0 points in mBO, when combined with the structure-aware loss. 
In contrast, applying the same loss to SlotContrast yields only marginal improvements, highlighting the importance of learning clear semantic boundaries within a curriculum learning setup.

\renewcommand{\thesection}{J} 
\section{Analysis of 2D and 3D Structure Loss}
\begin{table}[h]
\centering
\small
\caption{\textbf{Comparison between spatio-temporal~(3D-SSIM) and spatial-only~(2D-SSIM) structure-aware loss on YouTube-VIS.}
}
\label{tab:2dssim}
\renewcommand{\arraystretch}{1}
\setlength{\tabcolsep}{18.3pt}
\begin{tabular}{l cc}
\toprule
\multirow{2}{*}{\textbf{SSIM Objective}} &
\multicolumn{2}{c}{\textbf{YouTube‑VIS}} \\ 
\cmidrule(lr){2-3}
& FG‑ARI\,$\uparrow$ & mBO\,$\uparrow$ \\ 
\midrule
\xmark \;\;\; SSIM & 42.9 & 33.8 \\ \hline
+2D-SSIM & 44.2 & 34.3 \\
\rowcolor{gray!20}
+3D-SSIM & 44.8 & 35.5 \\
\bottomrule
\end{tabular}
\end{table}

\begin{table*}[t]
\centering
\small
\caption{Results with varying number of slots on MOVi-C.}
\label{tab:robustness}
\renewcommand{\arraystretch}{1.0}  
\setlength{\tabcolsep}{19.5pt} 
\begin{tabular}{l cc cc cc}
\toprule
\multirow{2}{*}{\textbf{Method}} &
\multicolumn{2}{c}{\textbf{slot=7}} &
\multicolumn{2}{c}{\textbf{slot=11}} &
\multicolumn{2}{c}{\textbf{slot=15}} \\
\cmidrule(lr){2-3}\cmidrule(lr){4-5}\cmidrule(lr){6-7}
& FG‑ARI\,$\uparrow$ & mBO\,$\uparrow$
& FG‑ARI\,$\uparrow$ & mBO\,$\uparrow$
& FG‑ARI\,$\uparrow$ & mBO\,$\uparrow$ \\ 
\midrule
SContrast & 74.9 & \textbf{27.9} & 69.3 & 32.7 & 61.8 & 31.2 \\
\rowcolor{gray!20}
Ours & \textbf{78.7} & 27.8 & \textbf{77.6} & \textbf{32.8} & \textbf{74.8} & \textbf{33.9} \\ 
\bottomrule
\end{tabular}
\end{table*}

Our framework's performance relies on sharpening semantic boundaries before slot expansion. 
We achieve this by complementing the standard MSE loss with a structure-aware SSIM objective. 
However, for video tasks, this structural loss can be applied in two ways: (1) as a 2D-SSIM, computed independently for each frame, or (2) as a 3D-SSIM, computed over spatio-temporal cubes.
In this work, we employed a 3D-SSIM objective since 2D-SSIM treats each video frame as an independent image, ignoring the temporal dimension. 
This objective lacks any incentive to maintain structural consistency across time, which is a critical signal for learning stable object representations across the temporal dimension.
To validate this design choice, we trained an identical model where our 3D-SSIM loss was replaced by a standard 2D-SSIM loss applied frame-by-frame. 
As shown in Tab.~\ref{tab:2dssim}, the 2D-SSIM variant suffers a performance decrease. 
This confirms our hypothesis that enforcing structural consistency across time is a key component of our method's success, ensuring that the sharpened boundaries learned in the early curriculum stages are temporally robust.

\begin{table*}[t]
\centering
\small
\caption{Experimental results on object dynamics prediction. SlotFormer~(SF) is used to evaluate each pretrained VOCL method.}
\label{tab:dynamics}
\renewcommand{\arraystretch}{1.0}  
\setlength{\tabcolsep}{17.9pt} 
\begin{tabular}{l cc cc cc}
\toprule
\multirow{2}{*}{\textbf{Method}} &
\multicolumn{2}{c}{\textbf{YouTube-VIS}} &
\multicolumn{2}{c}{\textbf{MOVi‑C}} &
\multicolumn{2}{c}{\textbf{MOVi‑E}} \\
\cmidrule(lr){2-3}\cmidrule(lr){4-5}\cmidrule(lr){6-7}
& FG‑ARI\,$\uparrow$ & mBO\,$\uparrow$
& FG‑ARI\,$\uparrow$ & mBO\,$\uparrow$
& FG‑ARI\,$\uparrow$ & mBO\,$\uparrow$ \\ 
\midrule
Recon & 27.4 & 28.9 & 50.7 & 25.9 & \textbf{70.6} & 24.3 \\ 
SContrast & 29.2 & 29.6 & 63.8 & \textbf{26.1} & 70.5 & \textbf{24.9} \\
\rowcolor{gray!20}
SlotCurri~(Ours) & \textbf{31.4} & \textbf{30.1} & \textbf{70.0} & 25.7 & 70.2 & 24.7 \\ 
\bottomrule
\end{tabular}
\end{table*}

\renewcommand{\thesection}{K} 
\section{Robustness to Varying Slot Numbers}
\label{sec:robustness_slots}
In unsupervised object-centric learning, the exact number of objects in a scene is typically unknown a priori.
Consequently, models are often deployed with a predefined number of slots that exceeds the actual object count to ensure all entities are captured.
However, this overestimation of slot capacity frequently leads to a degradation in performance for conventional methods, as excess slots tend to fragment single objects into multiple parts.
To evaluate the robustness of our method against such capacity mismatches, we conduct experiments on the MOVi-C dataset by varying the number of slots $K \in \{7, 11, 15\}$.

The results are summarized in Tab.~\ref{tab:robustness}.
We observe that the baseline, SlotContrast, suffers from a severe performance drop as the slot capacity increases.
Specifically, when $K$ is increased from 11 to 15, its FG-ARI plummets by over 7 points (from 69.3 to 61.8), indicating that the model struggles to handle redundant slots and succumbs to over-fragmentation.
In contrast, our method demonstrates remarkable stability across all settings.
Even when the number of slots significantly overestimates the scene complexity~($K=15$), SlotCurri maintains a high FG-ARI of 74.8, outperforming the baseline by a large margin~(+13.0).
This confirms that our curriculum-based strategy effectively suppresses the activation of redundant slots, allowing the model to be robustly deployed without precise knowledge of the scene's object count.

\renewcommand{\thesection}{K} 
\section{Discussion on Failure Modes and Limitations}
While SlotCurri effectively mitigates over-fragmentation, we identify two primary areas for future investigation.

\textbf{Under-fragmentation of Small Objects.}
First, SlotCurri is less effective on datasets where under-fragmentation is the primary challenge.
As shown in Fig.~\ref{Fig.movie}, our predictions on the MOVi-E dataset, which contains a large number of small objects, can fail to delineate clear boundaries~(white circles in the first video) or leave small objects spatially entangled~(white circles in the second video).
We attribute this limitation to (1) the low spatial resolution~($24 \times 24$) of the feature maps, which inherently blurs fine-scale structures, and (2) the lack of explicit guidance for distinguishing between similar but spatially overlapping objects.
As future work, we plan to leverage overlapping image patches produced by processing the original frames along with the spatially-shifted frames.
By exploring the semantic differences between partially overlapping patches, we expect to better capture fine-grained structures, thereby alleviating under-fragmentation.

\textbf{Refining the Curriculum Schedule.}
Another limitation of SlotCurri is that it relies on a predefined curriculum, where the number of stages~($M$) and the iteration timings for slot expansion (\textit{e.g.}, at 10\% and 25\% of training) are set as hyperparameters. 
Our extensive experiments demonstrate that this scheduling strategy is robust and highly effective. 
However, as a fixed schedule, it may require minor manual tuning to achieve optimal performance on new datasets with significantly different characteristics. 
This presents a promising direction for future research: the development of scene-adaptive slot schedules, as noted in our conclusion. 
Such a mechanism could, for example, automatically trigger slot expansion based on metrics like the plateauing of reconstruction error, making the framework more robust and general.


\renewcommand{\thesection}{L} 
\section{SlotCurri on Object Dynamics Prediction}
\label{sec:dynamics}
Object dynamics prediction aims to forecast the future states of objects based on their historical observations, serving as a critical benchmark for evaluating the temporal consistency and physical meaningfulness of learned representations.
To validate the effectiveness of the slot representations acquired by SlotCurri in a practical downstream scenario, we conduct experiments on the object dynamics prediction task.
Following standard protocols~\cite{slotcontrast}, we train a dynamics predictor module~\cite{slotformer} on top of the frozen slots learned by our model.

The quantitative results are summarized in Tab.~\ref{tab:dynamics}.
In real-world scenarios, SlotCurri demonstrates a significant performance improvement over existing baselines, confirming that our curriculum-based approach yields representations that are far more robust to complex dynamics.
Notably, on the MOVi-C dataset, our method achieves a remarkably high FG-ARI compared to prior works, indicating that our slots successfully capture distinct object identities.
On the MOVi-E dataset, we observe a trend consistent with the VOCL benchmarks: since the primary challenge in MOVi-E stems from under-fragmentation, which differs from the primary target of SlotCurri, the performance gains are naturally less pronounced.
Nonetheless, our method demonstrates performance that remains competitive with the state-of-the-art.

\renewcommand{\thesection}{M}
\section{Additional Results on COCO Dataset}
\label{sec:coco_results}
Since SlotCurri operates as a curriculum learning framework that progressively expands the slot capacity, its applicability extends beyond video domains to static image object-centric learning.
To verify this generalizability, we conduct experiments on the COCO 2017 dataset.
For the static image setting, we adapt the SSIM loss from 3D to 2D and exclude the cyclic inference mechanism, as it relies on temporal dynamics.
Consequently, the model is trained using the reconstruction-guided slot curriculum combined with the 2D SSIM loss.

The quantitative results are presented in Tab.~\ref{tab:coco}.
Consistent with our observations in video benchmarks, SlotCurri effectively mitigates the over-fragmentation problem in the image domain.
This improvement is evidenced by the substantial increase in Image-ARI compared to the reconstruction-based baseline, demonstrating that our curriculum strategy successfully guides slots to capture coherent object instances even without temporal cues.

\begin{table}[t]
\centering
\small
\caption{\textbf{Quantitative evaluation on the MS COCO dataset.}
SlotCurri achieves a significant improvement in Image-ARI by effectively resolving the over-fragmentation issue inherent in the reconstruction-based baseline.}
\label{tab:coco}
\renewcommand{\arraystretch}{1}
\setlength{\tabcolsep}{11pt}
\begin{tabular}{l cc}
\toprule
\multirow{2}{*}{\textbf{Method}} &
\multicolumn{2}{c}{\textbf{COCO}} \\
\cmidrule(lr){2-3}
& Image-ARI\,$\uparrow$ & Image-mBO\,$\uparrow$ \\
\midrule
Reconstruction & 40.5 & 28.8 \\
\rowcolor{gray!20}
SlotCurri~(Ours) & 43.4 & 28.9 \\
\bottomrule
\end{tabular}
\end{table}

\renewcommand{\thesection}{N}   
\section{Additional Qualitative Results}
\begin{figure*}[t]
\centering
\includegraphics[width=1.\textwidth]{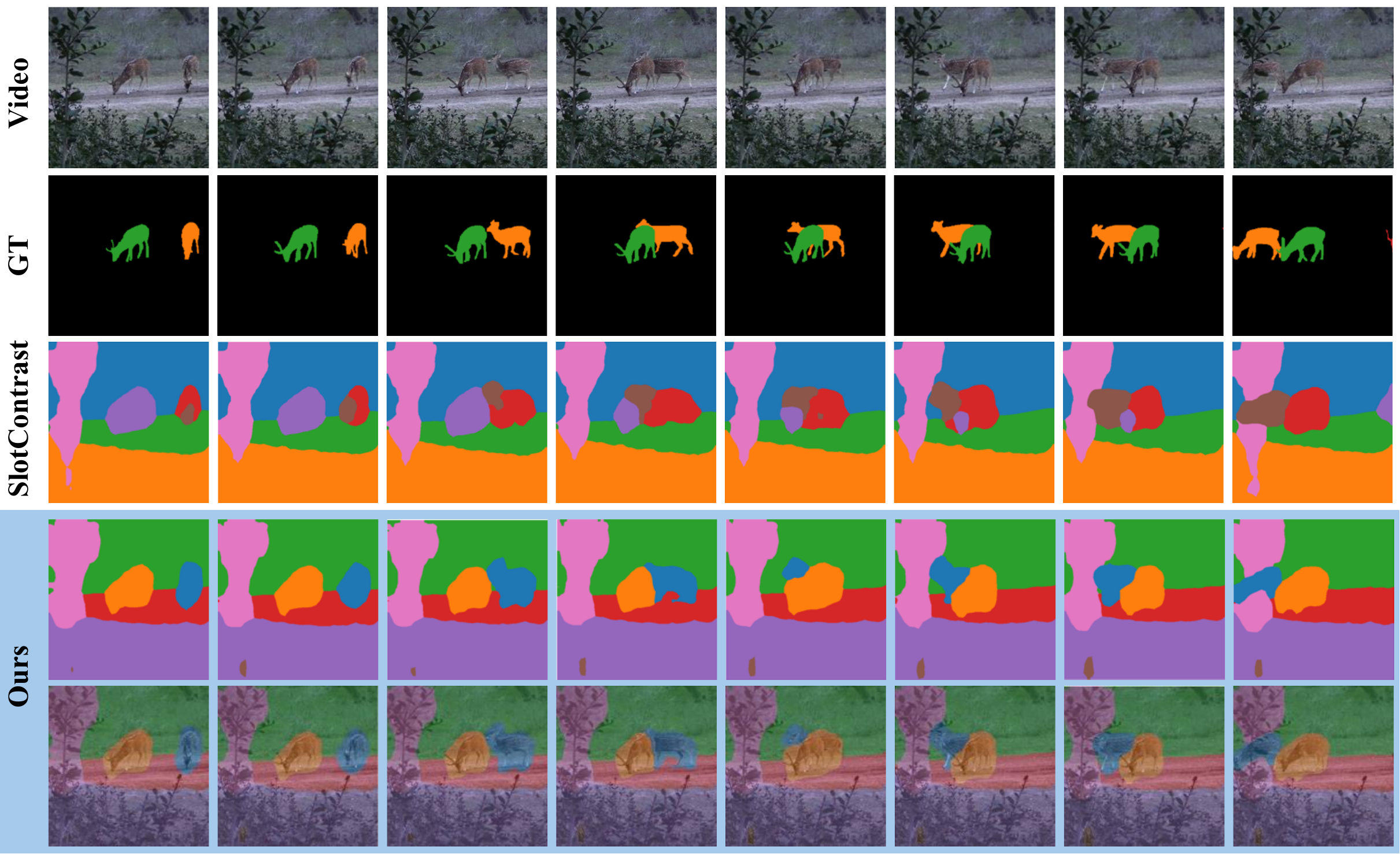} 
\includegraphics[width=1.\textwidth]{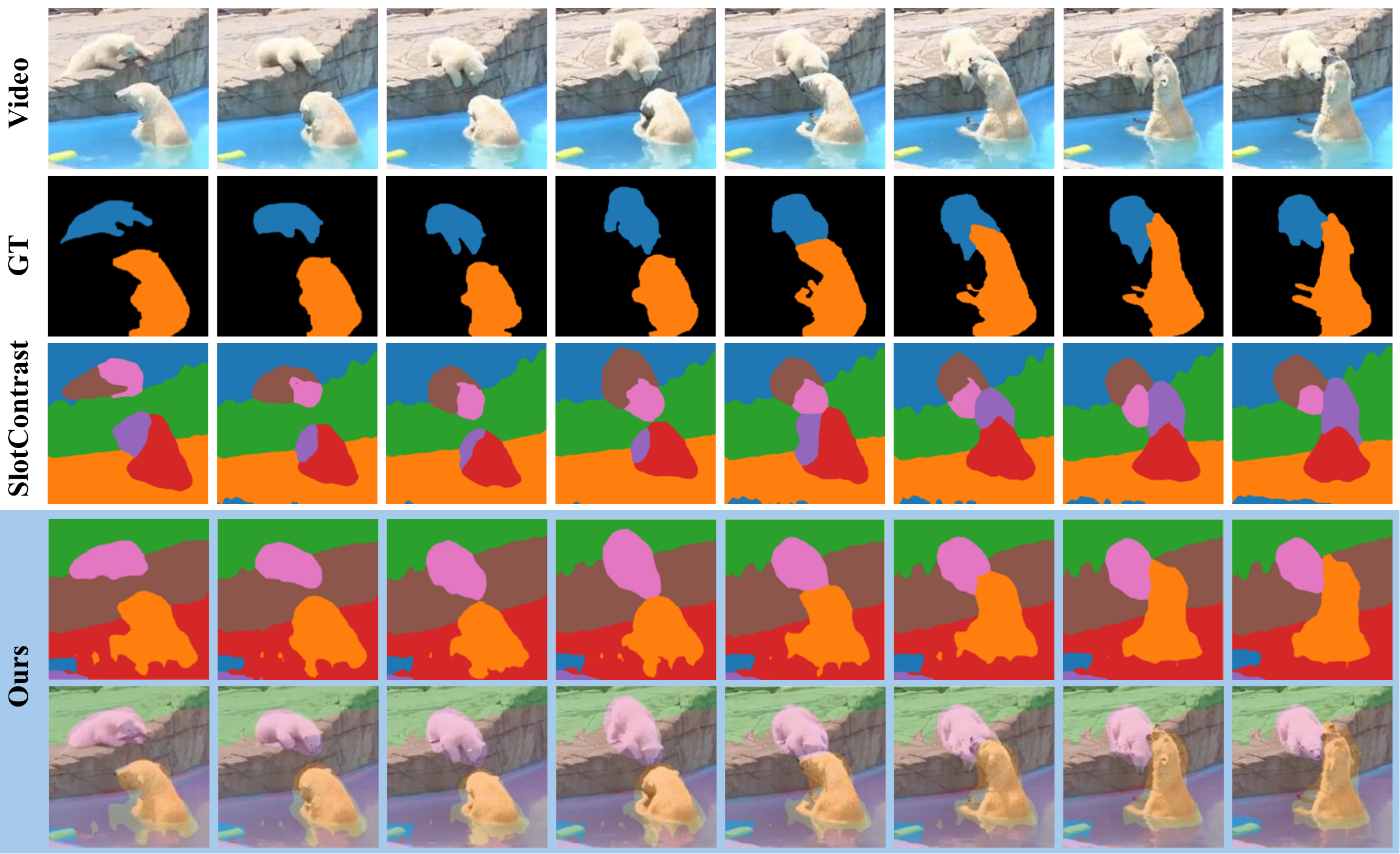} 
\caption{\textbf{Qualitative results on the YouTube-VIS dataset.} 
}
\label{Fig.ytvis_qual}
\end{figure*}

\begin{figure*}[t]
\centering
\includegraphics[width=1.\textwidth]{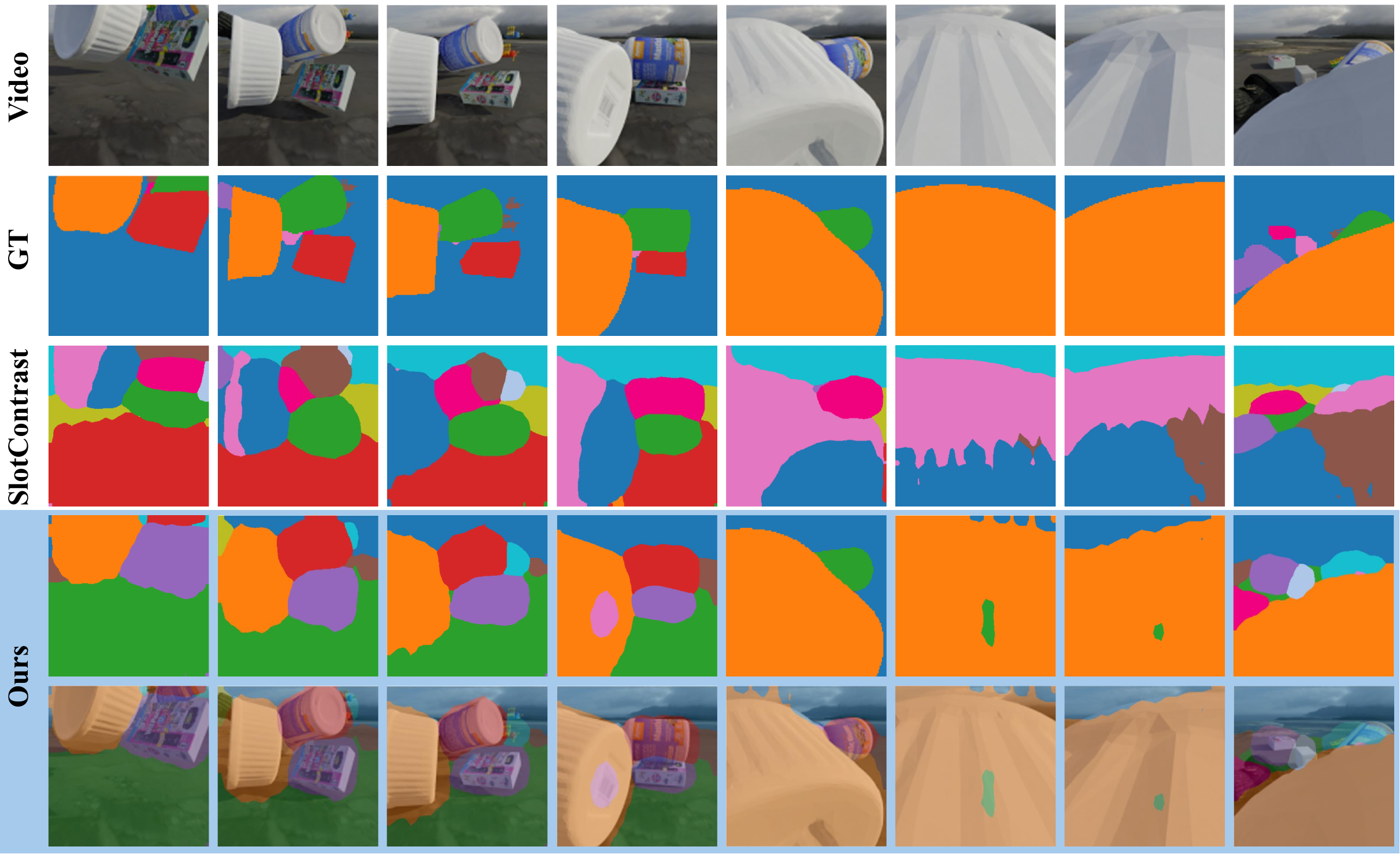} 
\includegraphics[width=1.\textwidth]{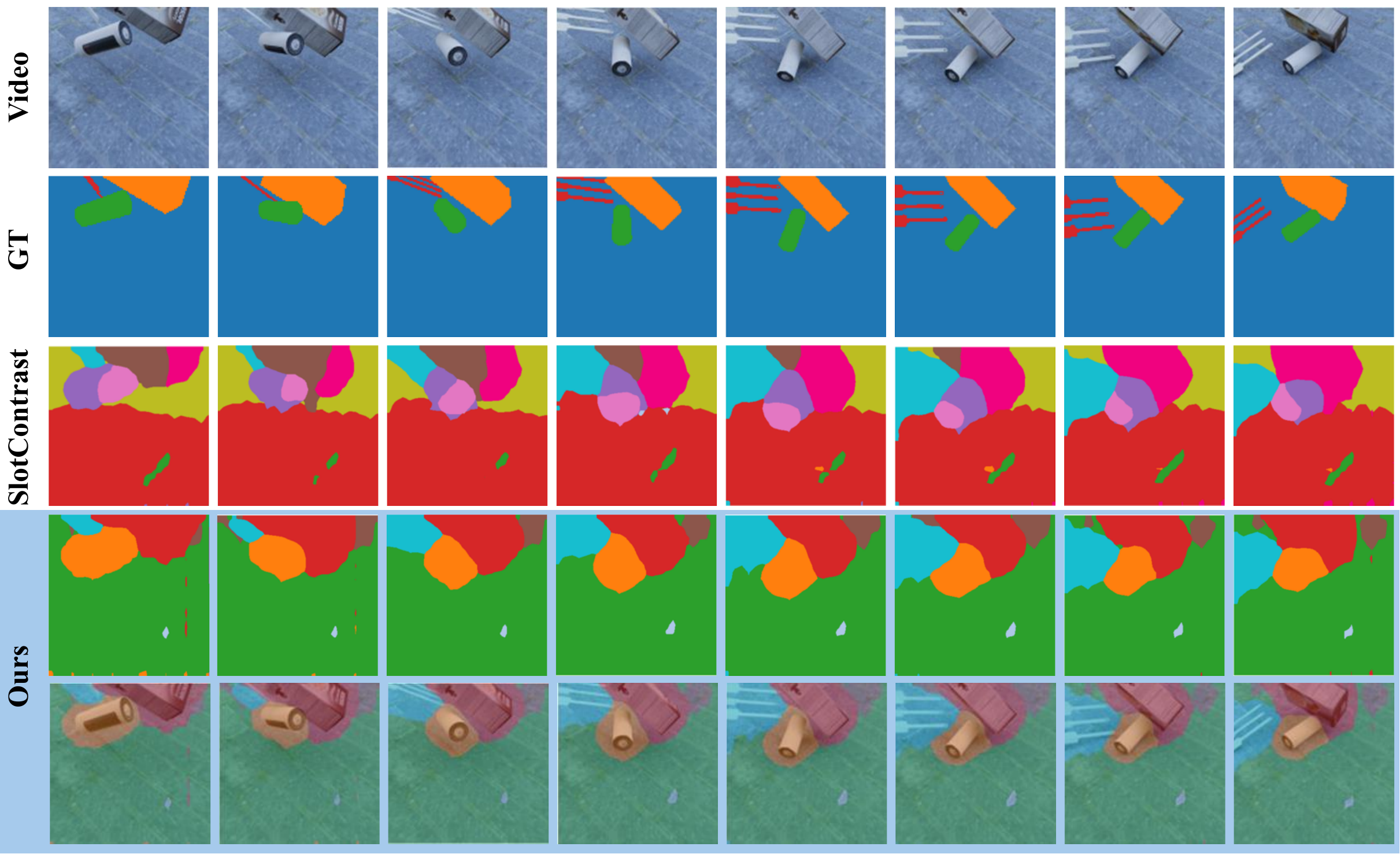} 
\caption{\textbf{Qualitative results on the MOVi-C dataset.} 
}
\label{Fig.movic_qual}
\end{figure*}

In Fig.~\ref{Fig.ytvis_qual} and \ref{Fig.movic_qual}, we present additional qualitative comparisons against SlotContrast~\cite{slotcontrast}.
Our method, SlotCurri, demonstrates strong object-level grouping, consistently assigning a single slot per semantic entity, even in an unsupervised setting.
In contrast, the SlotContrast exhibits clear signs of over-fragmentation.
For instance, in the first YouTube-VIS video, the deer~(masked with an orange mask in GT) is subdivided into red and brown masks in SlotContrast.
Similarly, in the second video, the body and head of bears are assigned to different slots, indicating spatial over-fragmentation.
These patterns persist across other videos and are also observed in the synthetic MOVi-C dataset.
While the baseline tends to over-segment coherent entities into multiple parts, SlotCurri maintains more consistent and compact object-level groupings across both space and time.

\newpage
{
    \small
    \bibliographystyle{unsrtnat}
    \bibliography{main}
}


\end{document}